\documentclass{article}
\usepackage[preprint]{neurips_2024}
\usepackage{natbib}
\setcitestyle{numbers,square}

\usepackage{natbib}
\setcitestyle{numbers,square}
\usepackage[utf8]{inputenc} % allow utf-8 input
\usepackage[T1]{fontenc}    % use 8-bit T1 fonts
\usepackage{hyperref}       % hyperlinks
\usepackage{url}            % simple URL typesetting
\usepackage{booktabs}       % professional-quality tables
\usepackage{amsfonts}       % blackboard math symbols
\usepackage{nicefrac}       % compact symbols for 1/2, etc.
\usepackage{microtype}      % microtypography
\usepackage{xcolor}         % colors
\usepackage{appendix}
\usepackage{subfigure}
\usepackage{multirow}
\usepackage{float}
\usepackage{bm}
\usepackage{diagbox}
\usepackage{setspace}
\usepackage{amsmath}
\usepackage{amssymb}
\usepackage{mathtools}
\usepackage{amsthm}
\usepackage{graphicx}
\usepackage{algpseudocode}
\usepackage{booktabs}
\usepackage{color}

\usepackage[T1]{fontenc}
\usepackage[utf8]{inputenc}
\usepackage{authblk}

\usepackage[linesnumbered,boxed,ruled,commentsnumbered]{algorithm2e}%%算法包，注意设置所需可选项

\theoremstyle{plain}
\newtheorem{theorem}{Theorem}[section]
\newtheorem{proposition}[theorem]{Proposition}

\theoremstyle{definition}
\newtheorem{definition}[theorem]{Definition}
\newtheorem{assumption}[theorem]{Assumption}
\theoremstyle{remark}

\newcommand{\Ex}{\mathbb{E}}
\newcommand{\Vx}{\mathbb{V}}
\newcommand{\R}{\mathbb{R}}

\newcommand{\bP}{\mathbb{P}}

\newcommand{\Popt}{\mathbb{P}^*}
\newcommand{\Pemp}{\hat{\mathbb{P}}}
\newcommand{\Pset}{\mathcal{P}}
\newcommand{\bz}{\boldsymbol{z}}
\newcommand{\bx}{\mathbf{x}}

\newcommand{\tz}{\Tilde{z}}

\newcommand{\tbz}{\Tilde{\bz}}
\newcommand{\hbz}{\hat{\bz}}
\newcommand{\bg}{\boldsymbol{g}}
\newcommand{\tbg}{\tilde{\bg}}
\newcommand{\hbg}{\hat{\bg}}
\newcommand{\bt}{\boldsymbol{\theta}}
\newcommand{\hR}{\hat{R}(\bt, \lambda)}
\newcommand{\tbx}{\tilde{\bx}}
\newcommand{\ty}{\tilde{y}}

\title{On the KL-Divergence-based Robust Satisficing Model}

\author[a]{Haojie Yan}
\author[b]{Minglong Zhou}
\author[c]{Jiayi Guo}

\affil[a]{Fudan University}
\affil[b]{Fudan University}
\affil[c]{corresponding author, Shanghai University of Finance and Economics}

\begin{document}

\maketitle

\begin{abstract}

Empirical risk minimization, a cornerstone in machine learning, is often hindered by the Optimizer’s Curse stemming from discrepancies between the empirical and true data-generating distributions. To address this challenge, the robust satisficing framework has emerged recently to mitigate ambiguity in the true distribution. Distinguished by its interpretable hyperparameter and enhanced performance guarantees, this approach has attracted increasing attention from academia. However, its applicability in tackling general machine learning problems, notably deep neural networks, remains largely unexplored due to the computational challenges in solving this model efficiently across general loss functions. In this study, we delve into the Kullback–Leibler-divergence-based robust satisficing model under a general loss function, presenting analytical interpretations, diverse performance guarantees, efficient and stable numerical methods, convergence analysis, and an extension tailored for hierarchical data structures. Through extensive numerical experiments across three distinct machine learning tasks, we demonstrate the superior performance of our model compared to state-of-the-art benchmarks.

\end{abstract}

\section{Introduction}
%Machine Learning (ML) algorithms aim to learn a model from training samples that can perform well on unseen test samples. Many classical algorithms rely on the Empirical Risk Minimization (ERM) framework,

Stochastic optimization models have been highly successful and extensively widely employed across a variety of domains of machine learning. A conventional approach for solving these models rely on the framework of Empirical Risk Minimization (ERM), which focuses on  
optimizing the expected performance over the empirical distribution derived from the training data.
%ERM is an empirical optimization approach that uses an empirical distribution to approximate the true data-generating distribution. 
The performance of the ERM model is greatly influenced by the quality of the training data in accurately representing the true distribution. For example, models trained based on ERM may perform poorly under a shifted distribution, sometimes even worse than an untrained model \cite{arjovsky2019invariant, creager2021environment, duchi2021learning, 
liu2021heterogeneous}, because the training samples does not accurately reflect the testing distribution. Even the train samples and test samples are generated from the same distribution, the empirical optimization results tend to be optimistically biased, leading to unsatisfactory performance on unseen samples. This phenomenon is usually called the Optimizer's Curse \cite{smith2006optimizer}.

To enhance model robustness towards discrepancy between the training data and the true data-generating distribution, distributionally robust optimization (DRO) is proposed and adopted across a variety of domains \cite{delage_distributionally_2010, wiesemann2014distributionally, bertsimas2018data,
mohajerin_esfahani_data-driven_2018-1}. The foundation of DRO models features an ambiguity set of possible true distributions and optimizes the worst-case expected performance over this set. Accounting for various possible distribution shifts, the out-of-sample performance is guaranteed to be as strong as that of the DRO objective if the true distribution lies in the ambiguity set. 

Recently, a novel framework, namely, Robust Satisficing (RS) is proposed to tackle the challenges of distribution shifts \cite{long_robust_2023, yang_fragility_2023}. Compared with DRO, RS protects the model beyond distribution shifts that are confined within an ambiguity set. Specifically, it aims to maintain an acceptable level of loss under a reference distribution and minimizes the fragility to impacts of distribution shifts on amplifying losses \cite{sim2021new}. 
%It has been shown that the RS model is closely related to regularization in linear regression and classification \cite{sim2021new}. 
The impact of distribution shifts is typically characterized by a probability distance function (e.g. Wasserstein distance or Kullback-Leibler divergence) from the reference distribution. Existing research focuses on transforming the RS model with Wasserstein distance into a tractable mathematical programming form if the loss function exhibits linearity or concave-convex properties, applies it to specific problems in different fields\cite{long_robust_2023, yang_fragility_2023, sim2021new}.
%The Wasserstein distance is suitable for describing geometric perturbations of distributions, such as those represented by adversarial examples \cite{sinha2017certifying, goodfellow2014explaining}. 
Within the realm of general machine learning tasks, 
the Wasserstein distance is inadequate for addressing non-geometric perturbations and may be hindered by computational challenges in practice \cite{nietert2024outlier,wang_sinkhorn_2023}. Alternatively, the Kullback-Leibler (KL) divergence, a metric for quantifying the disparity in information between two distributions, is widely adopted in machine learning applications, such as label distribution shift \cite{garg2022domain, zhang2021coping, garg2023rlsbench}, 
disentangled representation learning \cite{kim2018disentangling, higgins2017beta}, domain generalization \cite{zhao2020domain, nguyen2022fedsr}. However, the performance of KL-divergence-based RS models has received limited investigation, with only one study addressing its application in binary classification \cite{yang_fragility_2023}, thereby its applicability in tackle general machine learning tasks, such as deep neural networks, remains unexplored.

In this paper, we establish a comprehensive analysis of the Kullback-Leibler Robust Satisficing (KL-RS) model, and provide necessary theoretical foundations. In addition, we explores the analytical interpretations behind this model, designs tailored solution algorithms, 
conducts thorough experiments,  and extends its applicability for hierarchical data structures.

\iffalse
Let alone when distributional shifts occur between the training and testing datasets，
commonly referred to as Out-of-Distribution Generalization (OOD).
Researchers such as \cite{arjovsky2019invariant, creager2021environment, duchi2021learning, liu2021heterogeneous, shen2020stable} have demonstrated the existence of performance gaps when models optimized via ERM are deployed in environments characterized by shifted distributions, leading to potentially unsatisfactory performance.
\fi

\iffalse
Furthermore, 
KL-RS can control the robustness of model smoothly by tuning 
the target performance. Unlike DRO, which selects distribution by adjusting 
the size, target performance in our KL-RS is a parameter that is easier to 
understand and tune.
\fi

\subsection{Contribution}
Our main contribution can be summarized as follows:

(\textit{i}). We introduce a novel KL-RS framework tailored for general machine learning tasks, which achieves the highest robustness given a tangible tolerance. In light of the hierarchical  nature of data generation in prevailing applications, we present an adaptation of KL-RS named hierarchical KL-RS.
%, which has not been done in existing literature. %The proposed framework protects model against distributional ambiguity within the entire support. 
%By minimizing the rate at which the model's performance deteriorates beyond the target performance, we obtain a model robust to distribution shifts.
%We also propose variations of KL-RS called Hierarchical KL-RS, which are tailored to certain data structural characteristic of prevailing machine learning tasks.

(\textit{ii}). As a comprehensive investigation of the KL-RS model, we establish interesting analytical interpretations and novel performance guarantees. Moreover, we develop numerical algorithms for solving both KL-RS and hierarchical KL-RS model based on alternative optimization and explorations of the monotonicity and convex structure, and provide analysis of convergence guarantees. Notably, we identify two pivotal advantages of our algorithm: unbiasedness and normalization, to ensure the efficient and stable performance compared with existing approaches.

%We establish the KL-RS model's analytical interpretation, desired properties, asymptotic guarantees, and advantages over DRO approaches. The analysis of the analytical interpretation and asymptotic guarantees are new in the literature. We propose an efficient solution method based on alternative optimization and a gradient descent algorithm, exploiting the convex model structure. The gradient estimates are unbiased for the KL-RS model, different from the biased estimates in tilted loss minimization. We provide convergence guarantees our solution algorithms.

%To optimize KL-RS, we decompose the parameters into two parts and propose an alternative optimization procedure to optimize them. We identified the convex structure of some parameters and designed an efficient search algorithm to solve this subset of variables. For the remaining parameters, we employed a biased gradient descent algorithm for solving. Furthermore, we provide convergence guarantees for this biased gradient descend algorithm.

(\textit{iii}). Finally, we conduct thorough numerical experiments across three distinct machine learning tasks (label distribution shift, long-tailed learning, fair PCA), and the (hierarchical) KL-RS model outperforms problem-specific state-of-the-art benchmarks and universal DRO benchmarks. Our results demonstrate the effectiveness and versatility of the (hierarchical) KL-RS model in practical machine learning applications. Additionally, we offer in-depth and visual analysis on how the tolerance and fragile measure can affect the performance under varying distribution shifts.

%We implement our algorithm in complex machine learning models including deep neural networks and conduct comprehensive experiments on classical datasets. We compared our algorithm with popular benchmarks, including DRO and state-of-the-art machine learning algorithms. The numerical performance substantiates that the proposed model outperform benchmarks across various metrics, justifying the practical relevance of the proposed KL-RS method.

\subsection{Related Work}
\iffalse
\textbf{Distributionally Robust Optimization} 

Distributionally Robust Optimization (DRO) is a novel problem formulation 
to deal with the situation where decision maker is faced with partial information, 
on the statistical properties of model parameters. The basic approach is to construct 
distribution ambiguity set using statistical 
distance or some statistical index. By optimizing performance on the 
worst case distribution, DRO provides performance guarantee
for the decision\cite{bertsimas_optimal_2005, delage_distributionally_2010,
hu2013kullback, mohajerin_esfahani_data-driven_2018-1,duchi_learning_2021, 
wang_sinkhorn_2023, nietert2024outlier}. The basic philosophy behind DRO 
is to optimize the worst case in the parameter distribution
ambiguity to provide performance guarantee for the decisions. 
In recent years, as a natural implementation of adversarial learning, DRO 
has garnered widespread attention within the machine learning community
\cite{netessine_wasserstein_2019}. It has been applied to the robust 
supervised machine learning\cite{chen2018robust}, Kalman filtering
\cite{shafieezadeh-abadeh_wasserstein_2018}, federated learning
\cite{mohri2019agnostic, deng2020distributionally, reisizadeh2020robust, 
zecchin2022communication}. The equivalence of DRO formulation and 
regularization has also been demonstrated.
\fi

Our paper centers on the development of the KL-RS model and establishes its wide applicability in machine learning tasks. Existing research on RS in related applications typically adopts the Wasserstein distance \cite{sim2021new, yang_fragility_2023}, because Wasserstein distance is suitable for capturing geometric perturbations of distributions, showcased in works on adversarial examples \cite{sinha2017certifying, goodfellow2014explaining}. However, it 
fails to measure non-geometric distribution shift which is common in machine learning community, e.g., transitional kernel shift in MDP \cite{ruan2023robust, wiesemann2013robust}, label distribution shift\cite{zhang2021coping, garg2022domain, garg2023rlsbench} and 
domain adaptation\cite{zhou2022domain, zhao2020domain}.
As the computational challenges encountered by Wasserstein-distance-based models under general loss function in practical applications, existing work typically focus on exploring tractable reformulations through assuming the linearity or concave-convex properties of the loss function \cite{long_robust_2023, sim2021new, yang_fragility_2023, sinha2017certifying}.

KL divergence has received significant attention in machine learning community in addressing distribution shifts \cite{zhao2020domain, kim2018disentangling, higgins2017beta}. Nevertheless, the performance of KL-RS models remains under explored nowadays, with a solitary study examining its performance in binary classification \cite{yang_fragility_2023}, indicating a notable gap in systematic investigation of this framework. The effectiveness of the KL-RS model has been validated in operations management applications \cite{zhou2022intraday, jaillet2022strategic, hall2015managing}. Most models are reformulated into conic optimization problems that can be solved via commercial solvers. These solution methods rely on the structure of the operations management applications and are not directly applicable to complex functions commonly seen in machine learning tasks, such as neural networks. 

\section{Preliminaries}
\label{sec: prel}

\textbf{Notation.} We denote vectors in boldface lowercase letters. We use $\tz$ and $\tbz$ with a tilde sign to denote a random variable and a random vector, respectively. A probability distributions is denoted by $\bP$, $\bP \in \Pset(\Omega)$, where $\Pset(\Omega)$ represents the set of all Borel probability distributions on the support $\Omega \subseteq \R^n$. Moreover, $\Ex_{\bP}[\tz]$ and $\mathbb{V}_{\bP}[\tz]$ are the expectation and variance of a random variable, $\tz \sim \bP$, over its distribution. Lastly, we use $[N]$ to denote the index set $\{1, 2, ..., N\}$.

\textbf{Kullback–Leibler divergence.} 
The Kullback–Leibler (KL) divergence, denoted as $D_{KL}(\mathbb{Q}\Vert \mathbb{P})$, is a measure of discrepancy of a probability distribution $\mathbb{Q}$ from a reference distribution $\bP$ as defined below:
\begin{equation} \label{Dist: KL}
  D_{KL}(\mathbb{Q}\Vert \mathbb{P})=
        \mathbb{E}_{\mathbb{Q}}\left[\log\left(\frac{d\mathbb{Q}}{d\mathbb{P}}\right)\right] \;\; \text{ if }\;\;\mathbb{Q}\ll\mathbb{P},
\end{equation}
where we use $\mathbb{Q}\ll\mathbb{P}$ to denote that $\mathbb{Q}$ is absolutely continuous with respect to $\bP$\cite{van2014renyi}. 
This non-symmetric measurement is closely related to concepts such as relative entropy, information divergence, and discrimination information. 

\textbf{Data-driven stochastic optimization.}
In the context of statistical learning, we want to minimize some prediction loss over all possible parameter values of the learning model. Let $\Theta$ denote the set of model parameters and measurable continuous function $l:\Theta \times \mathbb{R}^n \to \mathbb{R}$ denote the loss function. For any $\bt\in\Theta$ and $\bm{z}\in\mathbb{R}^n$, $l(\bt,\bm{z})$ measures the prediction error of model $\bt$ on a data realization $\bm{z}$. A stochastic optimization model typically solves the following problem:
 \begin{equation} \label{Prob: sa}
   \min_{\bt \in \Theta} \Ex_{\Popt} [l(\bt,\tbz)],
  \end{equation}
where $\Popt$ is the true data-generating distribution of $\tbz$. 

In practice, we typically do not know the true distribution $\Popt$. A common approximation is to replace the true distribution with the empirical distribution constructed from a set of historical samples of size $N$, $\{\hat{\bm{z}}_1, \ldots, \hat{\bm{z}}_N\}$. We use $\Pemp$ to represent the empirical distribution, i.e., $\Pemp(\tbz=\hbz_i)=1/N$. The empirical optimization problem is cast as follows:
\begin{equation} \label{Prob: eop}
   E_{0}=\min_{\bt \in \Theta} \Ex_{\Pemp} [l(\bt,\tbz)]   = \min_{\bt \in \Theta} \frac{1}{N} \sum_{i=1}^N l(\bt,\hbz_i).
  \end{equation} 

The empirical optimization problem \eqref{Prob: eop} is widely adopted in the field of machine learning. For instance, the Empirical Risk Minimization (ERM) framework falls within the scope of \eqref{Prob: eop}, defining a broad range of models and applications.

\textbf{Distributionally robust approaches.}
When a stochastic program is calibrated based on one dataset but is evaluated on a separate dataset, the out-of-sample performance is frequently found to be unsatisfactory. This phenomenon is known as the \textit{optimizer’s curse} \cite{smith2006optimizer}. Inherently, the empirical optimization may yield an inferior solution that performs poorly out-of-sample when the historical data is not sufficient \cite{kuhn2019wasserstein, shapiro2021lectures}.

To account for the discrepancy between the empirical distribution and the true data-generating distribution, Data-driven Distributionally Robust Optimization (DRO) incorporates a probability-distance-based ambiguity set of potential distributions $\mathcal{B}(r)$, and solves
\begin{equation} \label{Prob: dro}
\min_{\bt \in \Theta} \max_{\bP \in \mathcal{B}(r)}\Ex_{\bP} [l(\bt,\tbz)]   \quad \text{where} \;\; \mathcal{B}(r)\triangleq\{\bP \in \Pset(\Omega): D(\bP\Vert\Pemp) \leq r\}.
\end{equation}
Common choices of probability distance $D$ include  Wasserstein metric, KL divergence, and generalized $\phi$-divergence \cite{mohajerin2018data, gao2023distributionally, duchi2021learning, duchi2023distributionally}. 

The classic DRO model hedges against $\Popt \in \mathcal{B}(r)$ but it leads to no guarantee when $\Popt$ is outside $\mathcal{B}(r)$. Recently, a novel development in optimization under uncertainty is the \textit{Robust Satisficing} framework \citep{long_robust_2023}, which can provide performance guarantees under all possible only distributions rather than distributions in $\mathcal{B}(r)$. The robust satisficing model can be written as the following:
\begin{equation}\label{Prob: rs}
\begin{aligned}
 \min_{\lambda\geq 0, \bt\in\Theta} & \quad \lambda\\
  \text{s.t.} \quad & \quad \mathbb{E}_{\mathbb{\bP}}[l(\bt,\tbz)]\leq \tau
  +\lambda D(\mathbb{\bP}\Vert \Pemp)  \quad \forall \bP \in \Pset(\Omega),
\end{aligned}
\end{equation}
where $D$ represents a probability distance measure and $\tau$ denotes a prescribed threshold for the prediction loss. The optimal value $\lambda^*$ is called the fragility measure \citep{long_robust_2023} or the fragility index \cite{yang_fragility_2023}, and it measures the robustness of the learning parameter $\bt$.

\section{KL Divergence based Robust Satisficing Model}
\label{sec: model}

\subsection{Problem Formulation and Guarantees}
\label{sec: for}
We adopt the KL Divergence in the robust satisficing model \eqref{Prob: rs}, and formulate a KL-Divergence-based Robust Satisficing (KL-RS) Model as follows:
\begin{equation}\label{Prob: klrs}
\begin{aligned}
 \min_{\lambda\geq 0, \bt \in\Theta} & \quad \lambda\\
  \text{s.t.} \quad & \quad \mathbb{E}_{\mathbb{\bP}}[l(\bt,\tbz)]\leq \tau 
  +\lambda D_{KL}(\mathbb{\bP}\Vert \Pemp)  \quad \forall \bP \in \{\mathbb{P}_0 \in \Pset(\Omega): \mathbb{P}_0\ll
  \Pemp\}.
\end{aligned}
\end{equation}
The decision variable $\lambda$ quantifies the maximally allowed magnitude of constraint violation caused by the discrepancy of the true distribution $\Popt$ from the empirical distribution $\Pemp$. Specifically, a small value of $\lambda$ suggests that only minor violations are expected even if the true distribution is significantly different from the empirical estimate.

%Besides, in contrast to sizing the ambiguity sets in \eqref{Prob: dro}, $T$ is a tolerance of optimality relative to the empirical model, which can be interpreted as the controller of the robustness.

According to the properties of the KL divergence, we can reformulate the KL-RS model \eqref{Prob: rs}.
\begin{theorem} \label{thm: eq}
 The KL-RS model \eqref{Prob: klrs} is equivalent to 
  \begin{equation}\label{Prob: klrs Sim}
    \begin{aligned}
 \min_{\lambda\geq 0, \bt\in\Theta} & \quad \lambda\\
  \text{s.t.} \quad & \quad \hR \leq \tau,
    \end{aligned}
    \end{equation}
with $\hR\triangleq\lambda\log\left(\mathbb{E}_{\hat{\mathbb{P}}}
        \left[\exp\left(l(\bt, \tbz)/\lambda\right)
        \right]\right)$.    
\end{theorem}
The term $\hR$ is called \textit{$\frac{1}{\lambda}$-tilted loss} in \cite{li2023tilted, li2020tilted, qi2022attentional}. Instead of selecting $\lambda$ as a hyperparameter for tilted empirical risk minimization, the KL-RS model selects hyperparameter $\tau$, an acceptable loss of optimality relative to the empirical risk model \eqref{Prob: eop}, and obtains the highest robustness by optimizing $\lambda$. The benefits of the KL-RS model include a more tangible selection of tolerance $\tau$ \cite{ruan2023robust} and enhanced numerical performance and stability, as demonstrated in
Section \ref{sec: alg} and Appendix \ref{sec: num kl-rs}.

This simplified formulation in Theorem~\ref{thm: eq} enables us better understand the KL-RS problem~\eqref{Prob: klrs}. We demonstrate the analytical interpretations of the KL-RS model and performance guarantees. For the sake of concise and clarity, all proofs of theorems and propositions are relegated to Appendix \ref{app: proof}.

\textbf{Analytical Interpretations.} 
We offer analytical interpretations of the KL-RS model from dual perspectives. From a model formulation standpoint, we demonstrate its characterization as an empirical mean-variance constrained optimization paradigm, particularly when $\tau$ closely aligns with the empirical loss $\mathbb{E}_{\hat{\mathbb{P}}}[l(\bt, \tbz)]$. Concerning optimal distribution, the KL-RS model enhances the influence of samples with large losses to bolster its robustness. Detailed discussions and propositions are provided in the Appendix \ref{sec: int}.

\textbf{Tail probability guarantee.} Beyond analytical interpretation, we present a tail probability guarantee of the KL-RS model.
\begin{proposition}\label{prop: probability inequality}
Given a non-negative number $\alpha$, we have
$$\hat{\mathbb{P}}(l(\bt, \tbz)\geq \tau+\alpha)\leq \exp(-\alpha/\lambda)$$
for every feasible solution $(\bt, \lambda)$ of the KL-RS model.
\end{proposition}
From Proposition \ref{prop: probability inequality}, we can see that the probability of exceeding the tolerance $\tau$ decays exponentially. Moreover, the optimal $\lambda_{N}^*$ attained by the KL-RS model yields the most favorable exponential rate, as $\exp(-\alpha/\lambda)$ is decreasing in $\lambda$, consistent with the observation in previous literature \cite{zhou2022intraday,yang_fragility_2023}.

\textbf{Asymptotic performance guarantee.}  
Note that if $\Popt$ is a discrete distribution supported by $K$ points, then $2N \cdot D_{KL}(\Popt \Vert \Pemp) \to \chi^{2}_{K-1}$ as $N \to \infty$ \cite{pardo2018statistical}. Based on this property of asymptotic convergence, we establish the following theorem.
\begin{theorem}\label{thm: asymptotic discrete}
  Suppose that $\Popt$ is a discrete distribution supported by $k$ points. For optimal solution $(\bt^*_N, \lambda^*_N)$ of the KL-RS model and a given non-negative radius $r\geq 0$, we have
  \begin{equation}
    \mathbb{P}^N\left(\mathbb{E}_{\Popt}
    [l(\bt^*_N, \tbz))< \tau+\lambda^*_N r\right)\geq \chi^{2}_{K-1}(\Tilde{y}
    \leq 2Nr) \quad \text{as } \; N \to \infty,
  \end{equation}
  where we use  $\mathbb{P}^N$ to denote the distribution that 
governs the distribution of independent samples $\{\hbz_{i}\}_{i\in[N]}$ 
drawn from $\Popt$, and  $\Tilde{y}\sim \chi^{2}_{K-1}$ is a chi-squared distribution with degree of freedom $K-1$.
\end{theorem}

 Intuitively, this asymptotic result suggests that the optimal decision $(\bt^*_N, \lambda^*_N)$ of the KL-RS model with $N$ samples enjoys a high confidence that the expected loss under the true distribution would not exceed the threshold $\tau$ by $\lambda^*_N r$. In addition, as $\chi^{2}_{K-1}(\Tilde{y} \leq 2Nr) \geq 1 - \left(m\cdot \exp(1-m)\right)^{(K-1)/2}$ with $m=2Nr/(K-1)$ by the Chernoff inequality, we can estimate the required sample size for a desired precision.

\begin{theorem}\label{thm: asymptotic continuous}
  Suppose that $\Popt$ is a continuous distribution and a constant $C$ such that $|l(\bt,\tbz)|\leq C$ exists. For every feasible solution $(\bt^*_N, \lambda^*_N)$ of the KL-RS model and a given non-negative radius $r\geq 0$, we have
  \begin{equation}
    \mathbb{P}^N\left(\mathbb{E}_{\Popt}
    [l(\bt^*_N, \tbz))< \tau+\lambda^*_N r\right)\geq \sup_{K\geq 2}\chi^{2}_{K-1}\left(\Tilde{y}
    \leq 2Nr-\frac{2NC}{K\lambda} \right) \quad \text{as } \; N \to \infty,
  \end{equation}
  where we use  $\mathbb{P}^N$ to denote the distribution that governs the distribution of independent samples $\{\hbz_{i}\}_{i\in[N]}$ 
drawn from $\Popt$, and  $\Tilde{y}\sim \chi^{2}_{K-1}$ is a chi-squared distribution with degree of freedom $K-1$.
\end{theorem}

%\textcolor{red}{
%In addition, by the Lemma 1 in \cite{inglot2006asymptotic}, we can upper bound for the cumulative distribution function for $\chi_{K-1}^{2}$ by the following inequality
%\begin{proposition}
%\begin{equation}
%\chi^{2}_{K-1}(\tilde{y}\leq 2Nr)\geq 1-\frac{1}{\sqrt{\pi}}\frac{2Nr}{2Nr-K+3}\sqrt{\frac{(2Nr)^{(K-2)/2}(K-1)^{(K-2)}}{\exp(Nr)}},\quad K\geq 3, \quad 2Nr > K-3.
%\end{equation}
%\end{proposition}
%From the above inequality, we can see that the confidence level of the upper bound $\tau+\lambda_{N}^{\ast}r$ approaches 1 at an exponential rate $\exp(-N/2)$ as the sample size increases.
%}

%This result is not only applicable to discrete distributions, but also true for continuous ones through the technique of discretization (See Appendix \ref{sec: asy conti}).  

%  Suppose $\mathbb{P}$ is a continuos distribution. 
%  Let $(\bt, \lambda)$ be a feasible solution of problem \eqref{2} 
%  and any non-negative $r\geq 0$, then under assumption \ref{blf} we have
%  \begin{equation}
 %     \lim_{N\to\infty}\mathbb{P}(\mathbb{E}_{\mathbb{P}}
 %     [l(\bt, \tbz)]\leq T +\lambda r)\geq 
 %     \sup_{k\in\mathbb{N}_{+}}\mathbf{P}(\chi^{2}_{k-1}\leq 2Nr
 %      -\frac{2NC_{0}}{k\lambda}),
 % \end{equation}
 % where $\mathbb{N}_{+}$ is the set of all positive integer numbers.
 % $\chi^{2}_{k-1}$ is a r.v. following $\chi^{2}$-distribution with degree of freedom $k-1$.
%\end{theorem}

\textbf{Finite-sample guarantee.} 
In the context of finite-sample guarantee, it is essential to recognize that the divergence measure $D_{KL}(\Popt\Vert\Pemp)$ can become unbounded if the supports of distributions $\Popt$ and $\Pemp$ differ. Mitigating this issue often involves the conventional recourse of Laplace smoothing, yielding a smoothed distribution $\Pemp^l$.
Now, we propose the following theorem on  $\Pemp^l$. 
\begin{theorem}\label{thm: finite discrete}
Suppose that $\Popt$ is a discrete distribution supported by $K$ points, and $\Pemp^l$ is a Laplace smoothing of $\Pemp$ with $N$ samples. For optimal solution $(\bt^*_N, \lambda^*_N)$ of the KL-RS model on $\Pemp^l$ and a threshold $\delta>0$, we have
  \begin{equation*}
    \mathbb{P}^N\left(\mathbb{E}_{\Popt}
    [l(\bt^*_N, \tbz))< \tau+\lambda^*_N r\right)\geq 1-\delta
  \end{equation*}
The  where we use  $\mathbb{P}^N$ to denote the distribution that  governs the distribution of independent samples $\{\hbz_{i}\}_{i\in[N]}$ 
drawn from $\Popt$, and  $r=\mathbb{E}_{\mathbb{P}^{N}}[D_{KL}(\Popt\Vert\Pemp^{l})]+
    \frac{6\sqrt{K\log^{5}(4K/\delta)}+311}{N}+
    \frac{160K}{N^{3/2}}$.    
\end{theorem}
It is noteworthy that $r\rightarrow \mathbb{E}_{\mathbb{P}^{N}}[D_{KL}(\Popt\Vert\Pemp^{l})]>0$ as $N \rightarrow \infty$, as Laplace smoothing introduces bias by interpolating between the empirical distribution and a uniform prior.

\subsection{Hierarchical KL-RS}
% In many practical scenarios, data comes in different labels or classes. Specifically, many data can be seen as generated according to some hierarchical structures. Understanding the hierarchical structures allows for a more nuanced understanding and modeling of complex relationships, thereby enhancing model performance, fairness, and robustness across various application domains. 

In numerous practical scenarios, data are observed or considered to follow hierarchical structures. Leveraging these structures allows for the modeling of intricate relationships, thereby enhancing model performance, fairness, and robustness across various application domains. 

Formally, we use $\mathbb{P}$ to denote the joint distribution of $(\tbz, \tbg)$, $\mathbb{P}_{\tbg}$ and $\mathbb{P}_{\tbz\vert\tbg}$ to denote marginal distribution for $\tbg$ and conditional distribution for $\tbz\vert\tbg$, respectively. Therefore, a two-layer hierarchical structure generates sample data by generating the group information $\tbg$ from $\Popt_{\tbg}$ and covariate information from $\Popt_{\tbz \vert \tbg}$.

This two-layer hierarchical structure is commonly seen across various fields, where $\tbg$ typically denotes grouping classes in classification problems \cite{lin2017focal, 
ren2018learning}, sub-populations in fair machine learning \cite{duchi2021learning, 
hashimoto2018fairness}, sub-tasks in agnostic machine learning \cite{finn2017model, collins2020task}, labeled environment in invariant risk minimization \cite{arjovsky2019invariant}, or side information in contextual stochastic 
bilevel optimization \cite{hu2024contextual}. The hierarchical structure typically arises from the intrinsic nature of the problem (e.g. labels of the data), although it can also be intentionally constructed in other cases (e.g. labels by clustering).

%To analyze two-layer hierarchical structure in the KL-RS framework, we first recall the chain rule of the KL divergence. Specifically, we can decompose the joint distribution divergence $D(\mathbb{Q}\Vert \mathbb{P})$ into  marginal distribution divergence $D(\mathbb{Q}_{\tbg}\Vert\mathbb{P}_{\tbg})$ and conditional distribution divergence $D(\mathbb{Q}_{\tbz\vert \tbg}\Vert\mathbb{P}_{\tbz\vert \tbg})$.  

%\begin{proposition}{(Chain Rule of KL Divergence)}
%For any two joint distributions of $(\tbz,\tbg)$, $\mathbb{P}$ and $\mathbb{Q}$, 
%  \begin{equation*}
%    \begin{aligned}
%        D_{\mathcolor{red}{KL}}(\mathbb{Q}\Vert \mathbb{P}
%        )=D_{\mathcolor{red}{KL}}(\mathbb{Q}_{\tbg}       \Vert\mathbb{P}_{\tbg})+\mathbb{E}_{\mathbb{Q}_{\tbg}}
%        D_{\mathcolor{red}{KL}}(\mathbb{Q}_{\tbz\vert\tbg}
%        \Vert\mathbb{P}_{\tbz\vert\tbg}).
%    \end{aligned}
%  \end{equation*}
%\end{proposition}

%By the chain rule, we can distinguish the impact of label distribution and covariate distribution. 
To analyze two-layer hierarchical structure in the KL-RS framework, we propose the following \textit{hierarchical KL-RS} model based on the chain rule of the KL divergence:
\begin{equation}\label{eq: hierachicalKL-RS}
  \begin{aligned}
      \min_{\bt\in\Theta, \lambda_{1}\geq 0, \lambda_2 \geq 0} \quad & \lambda_{1}+w\lambda_{2}\\
      \text{s.t.} \quad & \mathbb{E}_{\mathbb{P}}\left[l(\bt, \tbz)\right]
      \leq \tau+\lambda_{1}D_{KL}(\mathbb{P}_{\tbg}\Vert \Pemp_{\tbg}) +\lambda_{2}\mathbb{E}_{\mathbb{P}_{\tbg}}D_{KL}(\mathbb{P}_{\tbz\vert\tbg}
      \Vert\Pemp_{\tbz\vert\tbg}), 
     \;\; \forall\mathbb{P}\ll\Pemp.
  \end{aligned}
\end{equation}
where $\lambda_{1}$ and $\lambda_{2}$ measures the fragility towards deviations in the marginal label distribution and conditional covariate distribution, and $w\geq 0$ is a weight hyperparameter. Similar to Theorem \ref{thm: eq}, we can reformulate the hierarchical KL-RS model \eqref{eq: hierachicalKL-RS} by the properties of KL divergence.
\begin{theorem}\label{thm: hierachicalquivalence}
 The hierarchical KL-RS model  \eqref{eq: hierachicalKL-RS} is equivalent to 
\begin{equation*}
  \begin{aligned}
  &\min_{\bt\in\Theta, \lambda_{1}\geq 0, \lambda_{2}\geq 0}\lambda_{1}+w\lambda_{2}\\
  &\lambda_{1}\log\left(\mathbb{E}_{\Pemp_{\tbg}}
  \left[ \exp\left(\lambda_{2}\log\left(
    \mathbb{E}_{\Pemp_{\tbz\vert\tbg}}
    \exp\left(l(\bt, \tbz)/\lambda_{2}\right)\right)/\lambda_{1}\right)\right]
  \right)\leq \tau.
  \end{aligned}
\end{equation*}
\end{theorem}
Once $\Pemp$ is given, the formulation in Theorem~\ref{thm: hierachicalquivalence} becomes a deterministic optimization model.

\section{Numerical Approaches} \label{sec: alg}

In this section, we provide numerical algorithms for solving the (Hierarchical) KL-RS model. The cornerstone of our algorithm design is the alternative optimization, a concept well-adopted in the literature  \cite{brown2009satisficing, zhou2022intraday}, with alternately optimizing $\bt$ and $\lambda$. 
%This concept of alternative optimization is well-adopted in the literature (see e.g. \cite{brown2009satisficing, zhang2019routing, zhou2022intraday}).

%Similar to solving RS problem in Operation Management, we adopt an alternative approach to optimize our KL-RS and Hierarchical KL-RS\cite{brown2009satisficing, zhang2019routing, zhou2022intraday}. Our optimization method alternates between $\bt$ and $\lambda$ where we fix one and optimize the other one. When $\bt$ is fixed, we can obtain optimal $\lambda$ by searching algorithm, 

\subsection{The Algorithm for Solving the KL-RS Model} \label{subsec: alg KL}
We consider the scope of alternative optimization. Supposing that $\lambda$ is fixed, the task at hand is to answer whether $\lambda$ is a feasible solution to the KL-RS model \eqref{Prob: klrs Sim}. In other words, we must verify whether there exists $\bt$ such that $\hat{R}(\bt,\lambda) \leq \tau$. Note that 
$$\hat{R}(\bt,\lambda) = \lambda\log\left(\mathbb{E}_{\hat{\mathbb{P}}}
\left[\exp\left(l(\bt, \tbz)/\lambda\right) \right]\right) \leq \tau \quad \Leftrightarrow \quad \mathbb{E}_{\hat{\mathbb{P}}} \left[f(\bt,\tbz;\lambda) 
    \right]\leq 1$$
with $f(\bt,\tbz;\lambda)\triangleq \exp\left(\frac{l( 
\bt, \tilde{\bz})-\tau}{\lambda}\right)$. Thus, we propose Algorithm \ref{alg: feasbility KL} to verify the feasibility.
\IncMargin{1em} % 使得行号不向外突出 
\begin{algorithm}\label{alg: feasbility KL}
    \SetAlgoNoLine % 不要算法中的竖线
    \textbf{Input:} $\lambda$\\
    Apply stochastic optimization method to solve $\min \mathbb{E}_{\hat{\mathbb{P}}} \left[f(\bt,\tbz;\lambda) 
    \right]$; \\
    Let the output objective value be $f^*$;\\
    \textbf{Output:} Boolean($f^*\leq 1$)
    \caption{Feasibility of the KL-RS model ($\lambda$)}
\end{algorithm}
\DecMargin{1em}
Note that one can choose any preferred stochastic optimization algorithm (e.g. SGD, SAGA, Adam) to solve $\min_{\bt\in\Theta} \mathbb{E}_{\hat{\mathbb{P}}} \left[f(\bt,\tbz;\lambda) 
\right]$, and it enjoys the same computational complexity as the empirical risk minimization of $l$ due to the following proposition.
\begin{proposition} \label{prop: Lp KLRS}
Suppose that $l(\bt,\hbz)$ is bounded. For any given sample $\hat{z}$ and $\lambda>0$, if $l(\bt,\hbz)$ is Lipschitz continuous or Lipschitz smooth or convex or strongly convex, then $f(\bt,\hbz;\lambda)$ is Lipschitz continuous or Lipschitz smooth or convex or strongly convex.   
\end{proposition}

Our approach not only exhibits the desirable property of $f$, but also distinguishes from existing tilted empirical risk minimization literature \cite{li2023tilted, li2020tilted, qi2022attentional}. Unlike existing approaches that directly minimize $\hat{R}(\bt,\lambda)$ which may suffer from biased gradient estimators, our method stands out for the unbiasedness by minimizing $\mathbb{E}_{\hat{\mathbb{P}}}\left[f(\bt,\tbz;\lambda)
\right]$ and the normalization by considering $\tau$, leading to a consistently effective and stable performance (See Appendix \ref{sec: num kl-rs} for a detailed discussion). 

Back to the scope of alternative optimization, supposing that $\bt$ is fixed, we benefit from the favorable property of monotonicity, i.e.
$\hat{R}(\bt,\lambda)$ is a \textit{non-increasing} function with respective to  $\lambda$,
and develop a bisection method (Algorithm \ref{alg: bisec KL-RS}) to solve the KL-RS model. The monotonicity of $\hat{R}(\bt,\lambda)$ guarantees the convergence of the bisection algorithm in $O\left(\log(1/\epsilon)\right)$ iterations.

\IncMargin{1em} % 使得行号不向外突出 
\begin{algorithm}\label{alg: bisec KL-RS}
\SetAlgoNoLine % 不要算法中的竖线
\SetKwInOut{Input}{\textbf{Initialization}}\SetKwInOut{Output}{\textbf{Output}} % 替换关键词
\Input{ $\underline{\lambda}=0$, a positive value $\lambda_0$, a precision $\epsilon>0$\\}
\While{\textnormal{Algorithm \ref{alg: feasbility KL}($\lambda_0$) == False}}
{$\underline{\lambda}\leftarrow \lambda_0$, $\lambda_0 \leftarrow 2\lambda_0$ }
\hspace{5mm}$\overline{\lambda}= \lambda_0$\\
\While{ $\overline{\lambda} - \underline{\lambda} \geq \epsilon$}
    {$\lambda_{\text{mid}}=(\overline{\lambda}+\underline{\lambda} )/2$\\
    \If { \textnormal{Algorithm \ref{alg: feasbility KL}($\lambda_{\text{mid}}$)==True} }
    {$\overline{\lambda}= \lambda_{\text{mid}}$}   \Else{$\underline{\lambda}= \lambda_{\text{mid}}$}
}
\textbf{Output: } $\lambda_{\text{mid}}$
    \caption{Solve the KL-RS by bisection method}
\end{algorithm}
\DecMargin{1em}

\subsection{The Algorithm for Solving the Hierarchical KL-RS Model} \label{subsec: alg H-KL}
The process of solving the hierarchical KL-RS model is similar but slightly more complicated. When $(\lambda_1,\lambda_2)$ is fixed, we adopt and tailor the biased stochastic gradient method \cite{hu2020biased} to verify the feasibility. When $\bt$ is fixed, we search $(\lambda_1,\lambda_2)$ by a bisection and golden-ratio search method due to the monotonicity and convexity. We relegate the details to the Appendix \ref{sec: num h kl-rs}.

%\iffalse
%\begin{enumerate}
%    \item Specific structures implies the algorithm
%\begin{itemize}
%    \item monotone property
%    \item reformulation of the objective function
%\end{itemize}
%    \item Algorithm KLRS
%    \begin{itemize}
%        \item Properties (unbias, numerical stable) and Complexity
%        \item Reason: Lipschitz [App]
%    \end{itemize}
%    \item Extend to group KLRS (name) [App]
%    \begin{itemize}
%        \item gradient structure
%        \item Convergence
%        \item Why? Lipschitz property [App]
%    \end{itemize}
%\end{enumerate}
%\fi

\section{Experiments}
\label{sec: exp}
In this section, we apply our proposed KL-RS model to several real-world machine learning tasks, establishing the wide applicability of our method. In section 
\ref{sec: label distribution shift}, we consider problems with
label distribution shifts; in section \ref{sec: long tailed learning}, 
we investigate long tail distribution classification;
in section \ref{sec: fair pca}, 
we illustrate the benefits of the KL-RS model in addressing the fairness issue in machine learning tasks.

\subsection{Label Distribution Shift}
\label{sec: label distribution shift}

Label distribution shift is a common example in out-of-distribution problems. Inherently, the proposed KL-RS model is suitable in this context. The optimal learning parameter determined by the KL-RS model enjoys a profile of performance guarantees with respect to the expected prediction loss under all possible distributions, leading to satisfactory out-of-sample performance.

In this section, we focus on a binary classification task on the HIV-1 dataset \cite{blake1998uci, 
asuncion2007uci,duchi2019variance, 
li2023tilted}. We implemented widely used machine 
learning models to benchmark the performance of the KL-RS model, including 
ERM\cite{vapnik1999overview}, HINGE\cite{ren2018learning}, 
HRM\cite{leqi2019human}, CVaRDRO\cite{duchi2019variance}, 
FocaL (FL)\cite{lin2017focal}, and LearnReweight (LR)\cite{ren2018learning}. KL-RS0.10 and KL-RS0.50 denote our KL-RS model with $\tau=(1+0.1)E_0$ and $\tau=(1+0.5)E_0$ respectively. In this experiment, we adopt a linear model with a sigmoid layer. We obtain the optimal learning parameters of all models using the same training dataset. The optimal learning parameters of the various model are then evaluated on testing datasets. To investigate the impact of label distribution shift, we vary the test dataset by its KL divergence from the training dataset. We adopt multiple metrics to evaluate model performance including the accuracy on positive sample, overall accuracy, F1 score, MCC, and the 90-th quantile of the rank error \cite{narasimhan2013relationship, 
norton2019maximization, qi2021stochastic}. For brevity, we relegate the detailed experiment setting and the definitions of these performance metrics to Appendix 
\ref{app: label distribution shift}.

\begin{figure*}[ht] %这里使用的是强制位置，除非真的放不下，不然就是写在哪里图就放在哪里，不会乱动
  \centering  %图片全局居中
  \vspace{-0.35cm} %设置与上面正文的距离
  \subfigtopskip=2pt %设置子图与上面正文或别的内容的距离
  \subfigbottomskip=2pt %设置第二行子图与第一行子图的距离，即下面的头与上面的脚的距离
  \subfigcapskip=-5pt %设置子图与子标题之间的距离
  \subfigure{
    \label{test overall acc}
    \includegraphics[width=0.45\linewidth]{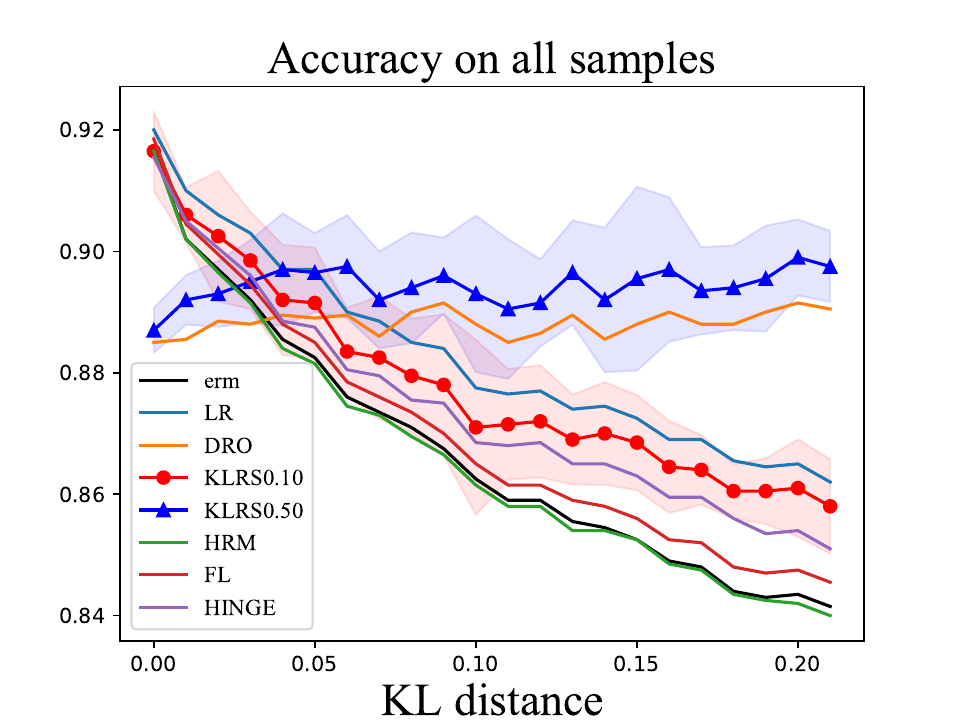}}
  \subfigure{
    \label{F1}
    \includegraphics[width=0.45\linewidth]{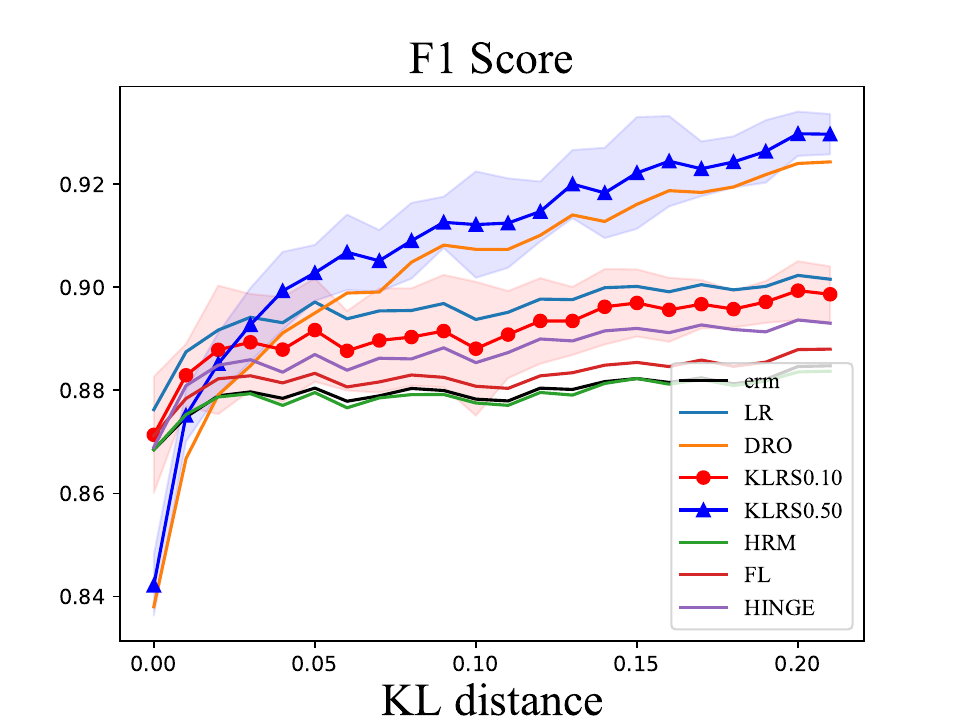}}
  \subfigure{
    \label{test MCC}
    \includegraphics[width=0.45\linewidth]{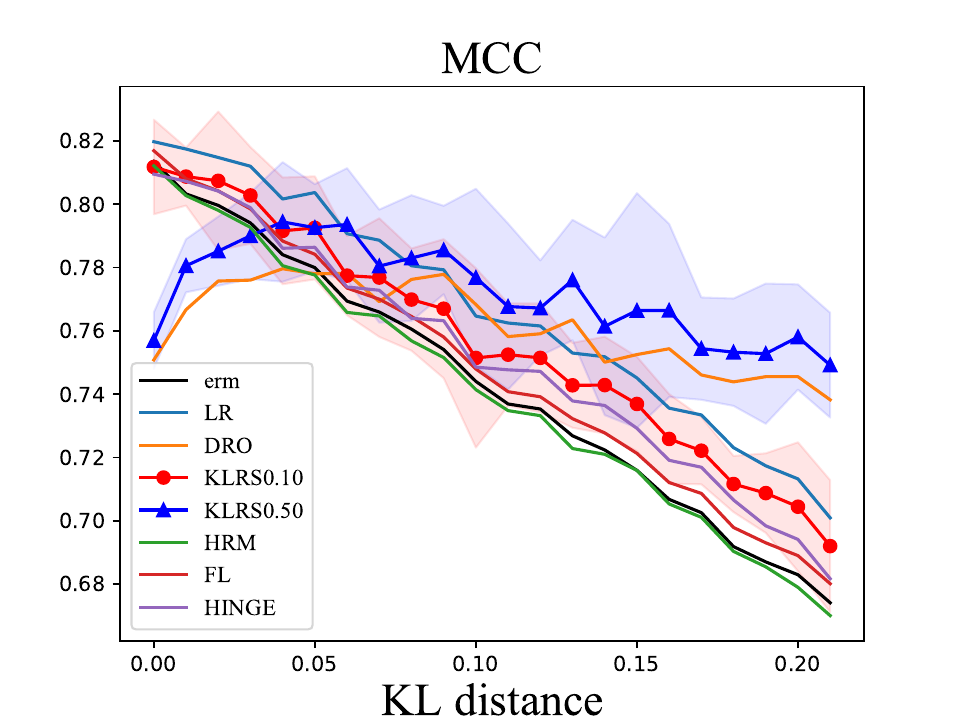}}
  \subfigure{
    \label{var90}
    \includegraphics[width=0.45\linewidth]{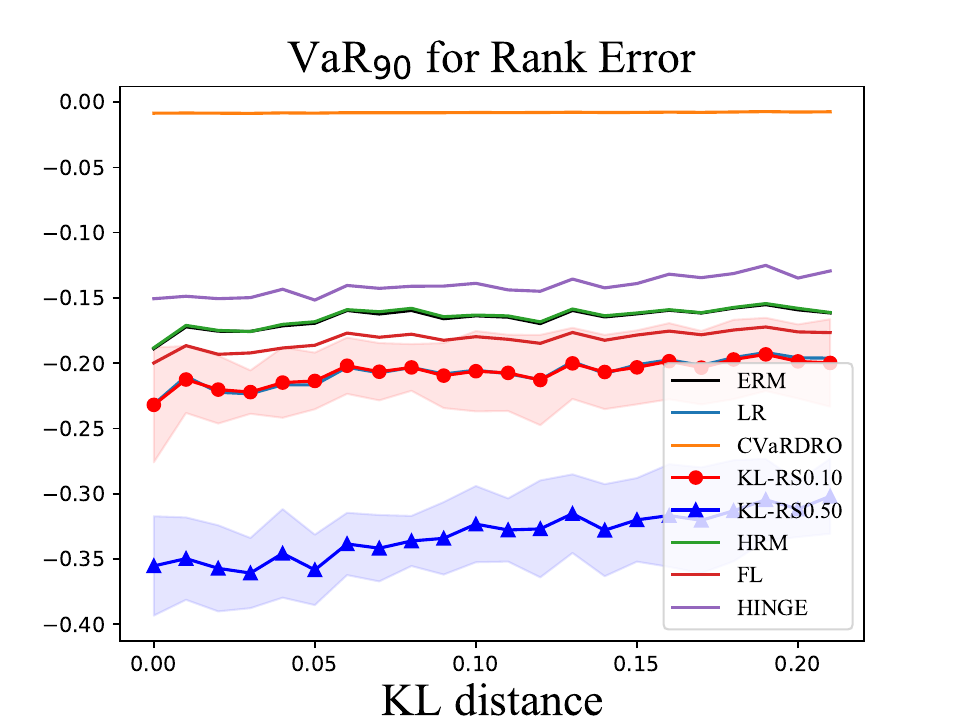}}
  \caption{Results of Label Distribution Shift Experiment}
  \label{pic: lds}
  \end{figure*}

We summarize the performance comparison in Figure \ref{pic: lds}. In each plot, the x-axis represents the KL divergence from the training dataset, and the y-axis records the value of a performance metric. For a binary classification model, higher accuracy, F1 score and MCC are desirable, and a smaller rank error is preferred. KL-RS0.10 slightly increases the model's robustness, and as the distribution shift gradually increases \emph{i.e.,} when the KL divergence of the testing dataset from the training dataset increases, various metrics show a significant decline, similar to ERM. However, our KL-RS0.10 dominates the performance of ERM across all distances and metrics. 
KL-RS0.50 significantly enhances the model's robustness, and as the distribution shift increases, there is no noticeable decline in these metrics. This is because KL-RS0.50 ensures uniform performance across both classes of data, leading to stronger out-of-sample generalization ability. However, this improvement comes at the cost of in-sample performance.

\iffalse
These figures demonstrate that our KL-RS model achieves better performance on various metrics than the benchmarks when facing label distribution shift, \emph{i.e.,} when the KL divergence of the testing dataset from the training dataset increases. The intuition is clear: the KL-RS model minimizes the fragility measure $\lambda$ and naturally leads to solutions that are robust against adversarial label distributions, rendering it superior than benchmarks as the KL divergence increases.  
\fi

\subsection{Long-Tailed Learning}
\label{sec: long tailed learning}

Long-tailed learning can be seen as a special case of distribution shift problems, where the class distribution in the training data is skewed while the the test class distribution is evenly distributed. Existing models that use average in-sample performance as the optimization criterion can be easily biased towards dominant classes and perform poorly on minority classes. Our KL-RS model can be integrated into existing methods and help alleviate this issue to enhance overall performance as well as performance in minority classes.

To account for distribution shift in class proportions, we adopt our Hierarchical KL-RS formulation in \eqref{eq: hierachicalKL-RS}. The group variable is specified by the class label in the data. For simplicity, we consider a baseline case where the proportionality factor in the objective function of Problem~\eqref{eq: hierachicalKL-RS} is given as $w=0$. This implicitly assumes that distribution shift only happens at the class proportion level while there exists no shift in the distribution of covariates. We conduct experiments on popular artificial datasets for long-tailed learning:
CIFAR10 (LT) and CIFAR100 (LT) \cite{cao2019learning}. These datasets are generated from CIFAR10 and CIFAR100 by Long-Tail strategy. The LT strategy uses a parameter $\rho$ to control the ratio between the size of the most common and most rare classes. The LT strategy performs downsampling on the samples under each class, resulting in a geometric progression of sample quantities across labels. We implement ERM \cite{vapnik1999overview}, Focal \cite{lin2017focal}, 
Ldam \cite{cao2019learning}, and CVaRDRO \cite{duchi2019variance} 
as our benchmarks. For brevity, we relegate the details of the experiment setting and benchmark models to Appendix 
\ref{app: long tailed learning}.

In this experiment, we use ResNet-32 as our model and present the performance comparison results in Table \ref{tab: lll}, where we include the standard deviation of the metrics in the brackets. Experiment results indicate that our KL-RS model can improve performance over ERM and CVaRDRO benchmarks. When combined with benchmark algorithms including Focal and Ldam models, our KL-RS model can lead to further improvements. For example, the KL-RS Ldam model achieves the highest average and worst-case accuracy in CIFAR10 (LT) dataset. Similarly, the best performing models in CIFAR100 (LT) dataset are KL-RS Focal and KL-RS Ldam. 

Additionally, the KL-RS model leads to more pronounced improvement over benchmarks when the long-tail distribution is more skewed, \emph{i.e.,} when $\rho$ becomes smaller. As we can observe from the results when the imbalanced factor is $\rho=0.01$, 
the improvement attained by the KL-RS Ldam model over other models is on average 14$\%$ on CIFAR10 (LT) and 13$\%$ on CIFAR100 (LT).

\begin{table*}[h]
  \centering
  \caption{Results of Long-Tailed Learning}
  \label{tab: lll}
  \begin{tabular}{|c|c|l|l|l|l|}
    \hline
    \multicolumn{1}{|c|}{\multirow{2}{*}{Dataset}}& $\rho$ &\multicolumn{2}{c|}{0.1}
    & \multicolumn{2}{c|}{0.01}\\
    \cline{2-6}
    & Algorithm & average acc & worst acc & average acc & worst acc\\
    \hline
    \multicolumn{1}{|c|}{\multirow{7}{*}{CIFAR10 (LT)}} &ERM& 74.93\scriptsize{(0.90)} & 65.77\scriptsize{(1.55)} & 52.01\scriptsize{(0.63)} & 14.72\scriptsize{(2.28)}\\
    \cline{2-6}
    &KL-RS& 76.28\scriptsize{(0.83)} & 66.43\scriptsize{(0.26)} & 61.80\scriptsize{(0.42)} & 50.32\scriptsize{(0.60)}\\
    \cline{2-6}
    &CVaRDRO& 75.25\scriptsize{(1.32)} & 66.83\scriptsize{(0.80)} & 53.66\scriptsize{(0.99)} & 10.46\scriptsize{(4.15)}\\
    \cline{2-6}
    &Focal& 73.55\scriptsize{(1.23)} & 62.34\scriptsize{(4.52)} & 50.83\scriptsize{(0.79)} & 14.35\scriptsize{(1.72)}\\
    \cline{2-6}
    &KL-RS Focal& 74.81\scriptsize{(0.70)} & 63.81\scriptsize{(2.05)} & 59.30\scriptsize{(1.36)} & 48.98\scriptsize{(0.51)}\\
    \cline{2-6}
    &Ldam& 81.86\scriptsize{(0.52)} & 71.60\scriptsize{(1.02)} & 59.61\scriptsize{(1.83)} & 9.46\scriptsize{(4.37)}\\
    \cline{2-6}
    &KL-RS Ldam& \textbf{82.87\scriptsize{(0.44)}} & \textbf{75.06\scriptsize{(1.17)}} & \textbf{70.68\scriptsize{(0.13)}} & \textbf{60.96\scriptsize{(1.99)}}\\
    \hline
    \multicolumn{1}{|c|}{\multirow{7}{*}{CIFAR100 (LT)}} &ERM& 40.28\scriptsize{(0.75)} & 1.98\scriptsize{(0.99)}& 24.84\scriptsize{(0.49)}& 0.00\scriptsize{(0.00)}\\
    \cline{2-6}
    &KL-RS& 41.86\scriptsize{(0.54)} & \textbf{15.18\scriptsize{(1.14)}} & 27.82\scriptsize{(0.63)} & 0.00\scriptsize{(0.00)}\\
    \cline{2-6}
    &CVaRDRO& 39.55\scriptsize{(2.34)} & 2.75\scriptsize{(1.06)} & 24.74\scriptsize{(1.53)} & 0.00\scriptsize{(0.00)}\\
    \cline{2-6}
    &Focal& 40.30\scriptsize{(0.49)} & 2.31\scriptsize{(1.14)} & 25.15\scriptsize{(0.56)} & 0.00\scriptsize{(0.00)}\\
    \cline{2-6}
    &KL-RS Focal& 40.67\scriptsize{(0.82)} & 13.53\scriptsize{(3.02)} & 27.69\scriptsize{(1.15)} & 0.00\scriptsize{(0.00)}\\
    \cline{2-6}
    &Ldam& 42.85\scriptsize{(1.14)} & 0.00\scriptsize{(0.00)} & 27.10\scriptsize{(1.03)} & 0.00\scriptsize{(0.00)}\\
    \cline{2-6}
    &KL-RS Ldam& \textbf{45.65\scriptsize{(0.79)}} & 11.55\scriptsize{(2.49)} & \textbf{31.44\scriptsize{(1.21)}} & 0.00\scriptsize{(0.00)}\\
    \hline
  \end{tabular}
\end{table*}

\subsection{Fair PCA}
\label{sec: fair pca}

Principal Component Analysis (PCA) is one of the most fundamental dimension reduction algorithm
in representation learning \cite{tantipongpipat2019multi}. In \cite{samadi2018price}, authors show that standard PCA will exaggerate the reconstruction error in one subpopulation over other subpopulations, which can have an equal size, in real-word dataset. Fair PCA is a novel method that aims to learn low dimensional representations and obtain uniform performance over all subpopulations \cite{tantipongpipat2019multi, 
kamani2022efficient}. The fair PCA model can be formulated as a minimax problem and relaxed into a semi-definite program which can be solved by off-the-shelf solvers.  

\begin{figure*}[ht] %这里使用的是强制位置，除非真的放不下，不然就是写在哪里图就放在哪里，不会乱动
  \centering  %图片全局居中
  \vspace{-0.35cm} %设置与上面正文的距离
  \subfigtopskip=2pt %设置子图与上面正文或别的内容的距离
  \subfigbottomskip=2pt %设置第二行子图与第一行子图的距离，即下面的头与上面的脚的距离
  \subfigcapskip=-5pt %设置子图与子标题之间的距离
  \subfigure[Performance of different subgroups]{
    \label{KL-RS_pca_loss}
    \includegraphics[width=0.32\linewidth]{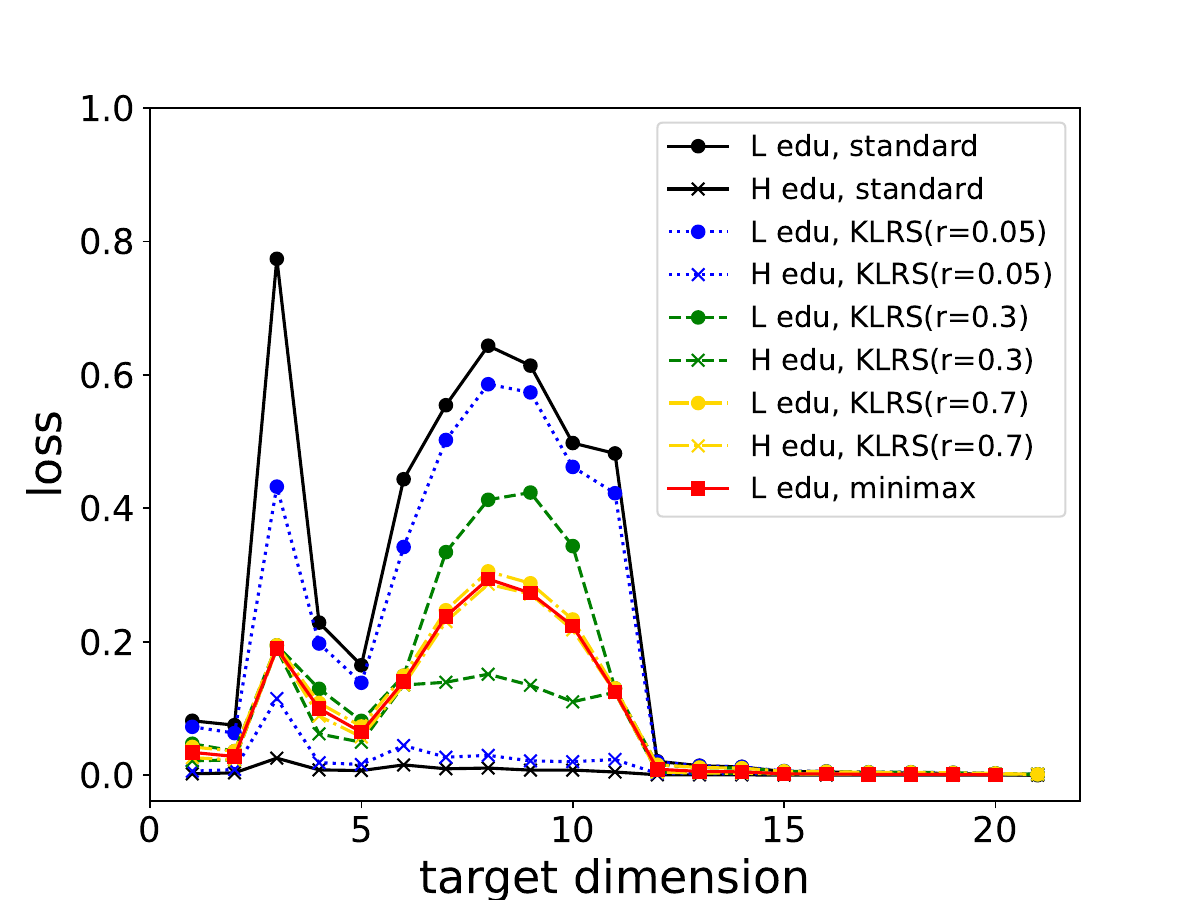}}
  \subfigure[Average performance]{
    \label{KL-RS_pca_average}
    \includegraphics[width=0.32\linewidth]{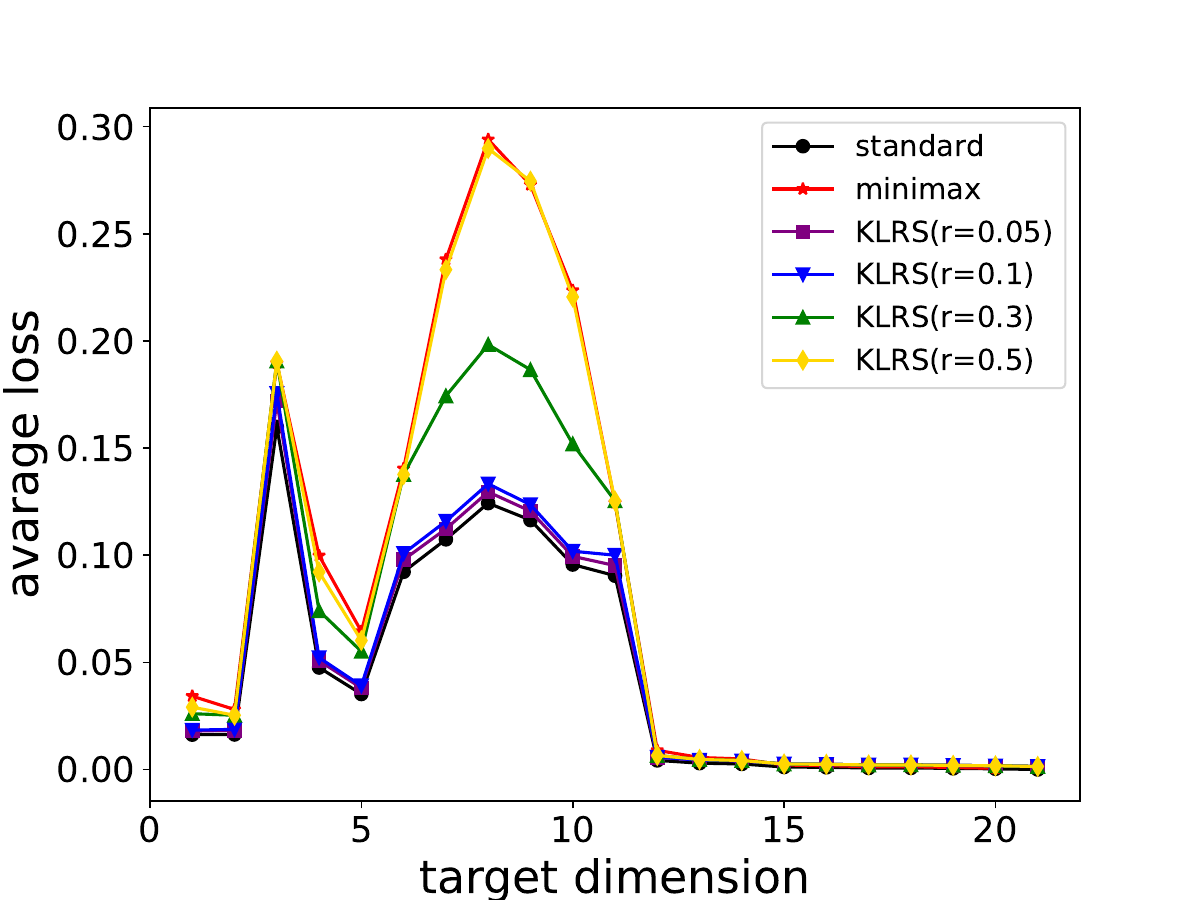}}
  \subfigure[Average vs. Difference]{
    \label{KL-RS_pca_frontier3}
    \includegraphics[width=0.32\linewidth]{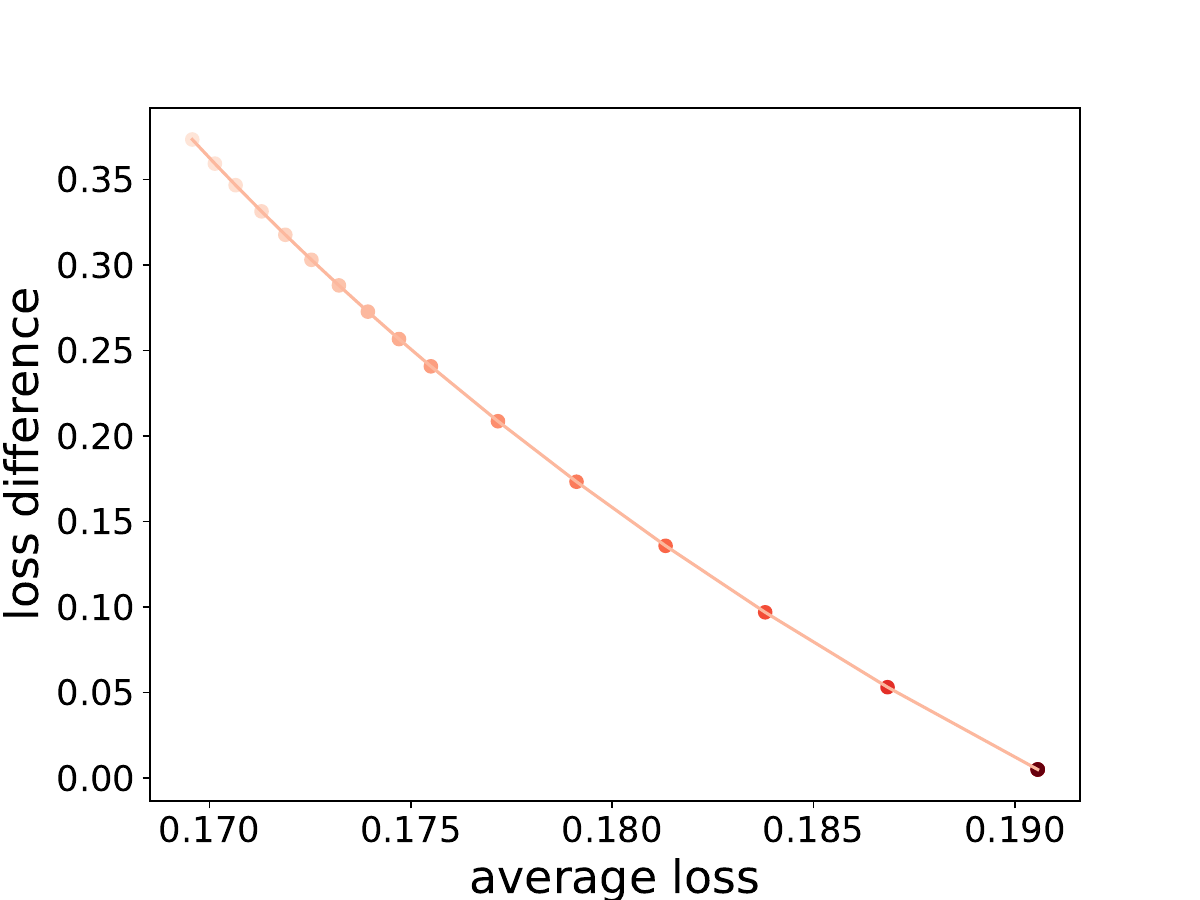}}
  \label{level}
  \caption{Fair PCA}
  \label{ref3}
  \end{figure*}

%However, a semi-definite program suffers from the curse of dimensions\cite{vandenberghe1996semidefinite}. 

As an alternative to the fair PCA model, we adopt our KL-RS paradigm to PCA tasks for balancing the performance difference over different subpopulation groups. Similar to \cite{samadi2018price}, we adopt reconstruction error as the loss function. For a matrix $Y\in\mathbb{R}^{a\times n}$, the loss incurred by projecting it to a rank-$d$ matrix $Z\in\mathbb{R}^{a\times n}$ is given by $\Vert Y-Z\Vert_{F}^{2}-\Vert Y-\hat{Y}\Vert_{F}^{2}$, where $\hat{Y}\in\mathbb{R}^{a\times n}$ is the optimal rank-d 
approximation of $Y$. Let $\bar{l}$ and $\underline{l}$ denote the maximum and minimum losses among of subgroups, respectively. The target parameter in the KL-RS model can be intuitively set as a convex combination of $\bar{l}$ and $\underline{l}$, \emph{i.e.,} $\tau = r \bar{l}+(1-r) \underline{l}, ~0\leq r\leq 1$. We can tune the hyperparameter $\tau$ conveniently by adjusting the parameter $r$. As the value of parameter $r$ increases, we
tend to obtain a projection matrix with more uniform performance across all subgroups. We implement KL-RS PCA with different parameter $r$ on Default Credit dataset
\cite{yeh2009comparisons}, which consists of two subgroups (high vs. low education). The benchmark models are standard PCA  and Fair PCA \cite{samadi2018price}. We relegate further details of the experiment setting to Appendix \ref{app: fair pca}.

We plot the  out-of-sample results in Figure \ref{ref3}. The x-axis represents different dimensions, \emph{i.e.,} the value of $n$. In Figure~\ref{KL-RS_pca_loss}, we plot the gap between the losses incurred in two subgroups under different models. We observe that the KL-RS model helps reducing this gap as $r$ increases. This is not attained for free. As we observe from Figure \ref{KL-RS_pca_average}, we may suffer from a higher average loss if we are overly concerned with ``fairness''. Nevertheless, we can obtain satisfactory average performance and fairness when $r$ is small. Such a trade-off is inherent in robust decision making. This is also discussed in fairness machine learning \cite{kleinberg2016inherent}. The 
advantage lies in our algorithm's ability to smoothly control this trade-off. 

\section{Conclusion}
We investigate a Kullback–Leibler-divergence-based robust satisficing model under a general loss function, extending the scope of existing literature in various aspects. Especially, this work makes several substantial contributions to the broader robust satisficing methodological framework in machine learning community, presenting new research directions and opportunities.

This paper still contains many areas worth further exploration. Firstly, extending from specific KL divergences to more general $\phi$-divergences is an avenue for future research. Simplifying and optimizing the dual forms of more general $\phi$-divergences are important points to investigate. Another aspect is exploring hierarchical structures. This paper only introduces hierarchical KL-RS, but utilizing such structures for modeling machine learning problems has not yet been explored.

Additionally, our research makes minimal assumptions about the true distribution $\Popt$, leading to relatively general conclusions. When our KL-RS framework is applied to specific problems, the true distribution may exhibit more definite characteristics, such as a normal distribution or a Bernoulli distribution. In these cases, the expression of our problem might be further simplified and presented in a more insightful mathematical form.
Exploring whether KL-RS can yield more insightful conclusions in more specific machine learning tasks compared to general cases might be a direction for future research.

%and design efficient solution algorithms tailored for complex machine learning tasks including deep neural networks.
%This results extends the scope of existing literature in various aspects, and  presenting new research directions and opportunities

%Through careful analysis, we present analytical interpretation, obtain convenient reformulation, and design efficient solution algorithms tailored for complex machine learning tasks including deep neural networks

\newpage
\bibliographystyle{plain}
\bibliography{reference}

\newpage
\appendix

\section{Analytical Interpretations} \label{sec: int} 

\textbf{Empirical mean-variance trade-off.} When the reference distribution is given by a normal distribution, the KL-RS model \eqref{Prob: klrs Sim} becomes an empirical mean-variance constrained model. As a concrete example, let us consider a linear loss $l(\bt,\tbz)=\bt^{\top}\tbz$ and $\tbz \sim \Pemp=N(\mu, \Sigma)$, and the constraint of the KL-RS model \eqref{Prob: klrs Sim} becomes
  \begin{equation*}
 \hR=\lambda\log\left(\mathbb{E}_{\hat{\mathbb{P}}}
        \left[\exp\left(l(\bt, \tbz)/\lambda\right)
        \right]\right) = \bt^{\top}\mu+
    \frac{\bt^{\top}\Sigma\bt}{2\lambda}  \leq \tau.
\end{equation*}
As $\bt^{\top}\mu$ and $\bt^{\top}\Sigma\bt$ are the mean and variance of $\bt^{\top}\tbz $, the KL-RS model robustify the solution by increasing the weight of the variance as much as possible while ensuring that the weighted sum of mean and variance is bounded below the tolerance $\tau$. In fact, this observation holds for any distribution of $\bm{\tilde{z}}$ if $\tau$ is a litter larger than $E_0$, as stated in the following proposition.

\begin{assumption}\label{be}
    $\Ex_{\Pemp}[l(\bt, \tbz)]\leq M_{1}, \forall \bt\in \Theta$.
\end{assumption}

\begin{assumption}\label{bv}
    $0<\epsilon\leq\Vx_{\Pemp}[l(\bt, \tbz)]\leq M_{2}, \forall \bt \in \Theta$. 
\end{assumption}

\begin{proposition}\label{smalltau}
  \label{largelambda} Under the Assumption \ref{be} and Assumption \ref{bv},
  for any $\bt\in\Theta$, we can choose $\tau =\mathbb{E}_{\hat{\mathbb{P}}}[l(\hat{\bt}_{N},
  \tbz)]+\frac{1}{b}\mathbb{V}_{\hat{\mathbb{P}}}[l(\hat{\bt}_{N}, 
  \tbz)]$ and $b> \max\left\{1, \frac{2e(M_{1}+M_{2})}{\epsilon}\mathbb{V}_{\hat{\mathbb{P}}}
  [l(\hat{\bt}_{N}, \tbz)]\right\}$ where $\hat{\bt}_{N}=\arg\min_{\bt}\mathbb{E}_{\Pemp}[l(\bt, \tbz)]$. 
  Then any feasible $\lambda$ of \eqref{Prob: klrs} must satisfy
  $\lambda\geq \frac{b\epsilon}{2e\mathbb{V}_{\hat{\mathbb{P}}}
  [l(\hat{\bt}_{N}, \tbz)]}$.
\end{proposition}

\begin{proposition}\label{prop: mv}
   If $\lambda$ is large enough, we have $\hat{R}(\bt, \lambda)
  =\mathbb{E}_{\hat{\mathbb{P}}}[l(\bt, \tbz)]+\frac{1}
  {2\lambda}\mathbb{V}_{\hat{\mathbb{P}}}[l(\bt, \tbz)]
  +o(\frac{1}{\lambda^{2}})$,
  where $\mathbb{V}_{\hat{\mathbb{P}}}[l(\bt, \tbz)]$ is the variance of $l(\bt, \tbz)$.
\end{proposition}
We note that the tolerance $\tau$ selected to marginally exceed empirical loss $\mathbb{E}_{\hat{\mathbb{P}}}[l(\bt, \tbz)]$ necessitates a significant large $\lambda$, thus transforming the KL-RS model into a mean-variance constrained optimization paradigm.

\textbf{Prioritization on large losses.} The KL-RS model magnifies the impact of samples with large losses (i.e., points that deviate significantly from the estimated values) as value of $\lambda$ decreases. For an illustrative example, we examine two extreme cases of $\hR$ as $\lambda \to 0^{+}$ and $\lambda \to \infty$ demonstrated below:
$$\displaystyle{\lim_{\lambda\to 0^{+}}}\hat{R}(\bt, \lambda)=
\max_{i\in[N]}l(\hbz_{i}, \bt), \quad  \displaystyle{\lim_{\lambda\to +\infty}}\hat{R}(\bt, \lambda).=\mathbb{E}_{\hat{\mathbb{P}}}[l(\tbz, \bt)]$$
As we see, when $\lambda$ approaches zero, the KL-RS model primarily addresses the extreme case. Conversely, as $\lambda$ tends towards infinity, the model puts an equal weight to all samples, reducing to an empirical optimization model \eqref{Prob: eop}.

Intuitively, when $\lambda$ decreases from $\infty$ to $0$ gradually, the KL-RS model distributes its attention from equally across all samples to the worst-case sample. In fact, we can quantify this effect by the following proposition. 
\begin{proposition} \label{prop: optDistn}
For a given $\lambda>0$, we obtain the optimal distribution $\mathbb{P}_0^*$ with
 $$\Tilde{\bz}_0^* \sim \mathbb{P}_0^*=\arg\sup_{\mathbb{P}\ll
  \Pemp}  \mathbb{E}_{\mathbb{\bP}}[l(\bt,\tbz)] - \lambda D_{KL}(\mathbb{\bP}\Vert \Pemp).$$  
 where $\mathbb{P}_0^*(\Tilde{\bz}_0^*=\hbz_i)=\frac{\exp(l(\bt, \hbz_{i})/\lambda)}
  {N \cdot \mathbb{E}_{\hat{\mathbb{P}}}[\exp(l(\bt, \tbz)/\lambda)]}$ for all $i \in [N]$.
\end{proposition}
We remark that the  constraint of KL-RS model in \eqref{Prob: klrs} can be reformulated as $\sup_{\mathbb{P}_0 \ll
  \Pemp}  \{\mathbb{E}_{\mathbb{\bP}}[l(\bt,\tbz)] - \lambda D_{KL}(\mathbb{\bP}\Vert \Pemp) \} \leq \tau$, and further simplified into $\mathbb{E}_{\mathbb{P}_0^*}[l(\bt,\tbz)] - \lambda D_{KL}(\mathbb{P}_0^* \Vert \Pemp) \leq \tau$ by Proposition \ref{prop: optDistn}. Now, we observe that the KL-RS model amplifies the influence of samples with large losses, where  $\mathbb{P}_0^*(\Tilde{\bz}_0^*=\hbz_i)$ serves as the relative weight attributed to each sample.

\subsection{An Illustrative Example}
\label{app: toy example}
In this subsection, we use a simple example to illustrate the above-mentioned interpretations.

\begin{figure*}[ht]
   %这里使用的是强制位置，除非真的放不下，不然就是写在哪里图就放在哪里，不会乱动
  \centering  %图片全局居中
  \vspace{-0.35cm} %设置与上面正文的距离
  \subfigtopskip=2pt %设置子图与上面正文或别的内容的距离
  \subfigbottomskip=2pt %设置第二行子图与第一行子图的距离，即下面的头与上面的脚的距离
  \subfigcapskip=-5pt %设置子图与子标题之间的距离
  \subfigure[]{
    \label{point estimation}
    \includegraphics[width=0.3\linewidth]{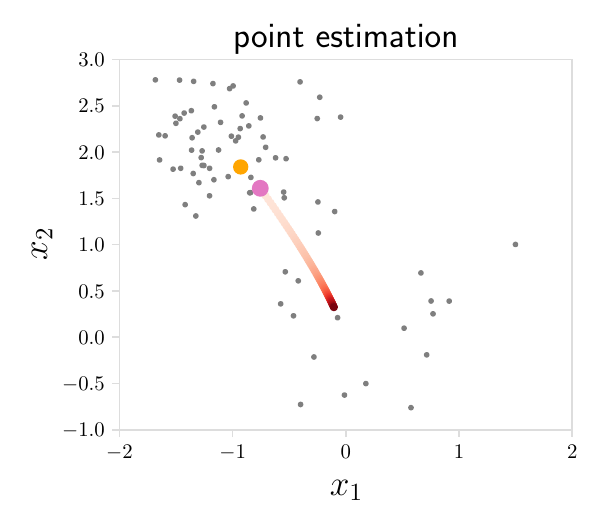}}
  \subfigure[]{
    \label{averagevsmax}
    \includegraphics[width=0.3\linewidth]{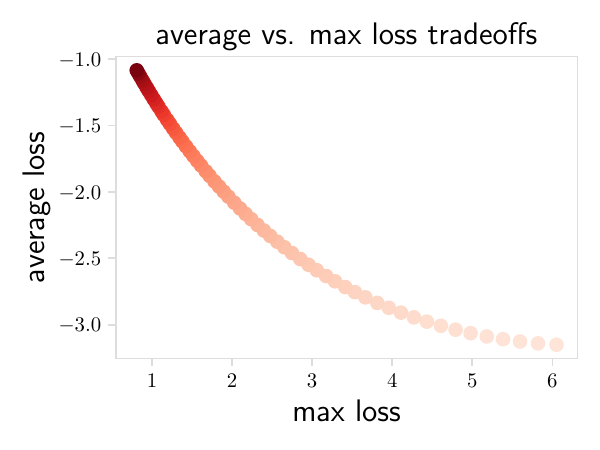}}
  \subfigure[]{
    \label{variancevstarget}
    \includegraphics[width=0.3\linewidth]{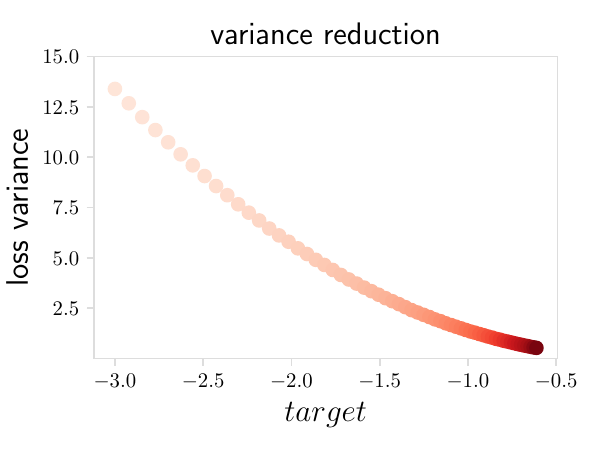}}
  \caption{Performance evaluations across varied $\tau$}
  \label{pic3}
\end{figure*}

Consider that 100 two-dimensional data points are generated from two distributions. In Figure \ref{point estimation}, 80 data points clustered in the top-left corner are sampled from a Normal distribution with mean $(-1,2)$ and variance $0.4 I$; the remaining 20 points clustered it the bottom-right corner are sampled from another Normal distribution with mean $(0.2,0.2)$ and variance $0.6 I$. In addition, we highlight the geometric mean of these points in pink and the arithmetic mean in yellow.

We conduct the KL-RS model training with loss function $l(\bt, \tbz)=\frac{1}{2}(\bt-\tbz)^2$ for the point estimation. As we vary the tolerance parameter $\tau$ from the empirical loss to infinity, the corresponding optimal solution $\bt^*(\tau)$ also varies. In Figure \ref{pic3}, points are shaded darker to denote $\bt^*(\tau)$ with larger $\tau$.

In Figure \ref{point estimation}, larger $\tau$ values lead to estimated points approaching minority class of points, consequently reducing maximum sample loss. This observation is also articulated in Figure \ref{averagevsmax}, where the average loss gradually increases while the maximum sample loss decreases. Figure \ref{variancevstarget} illustrates a reduction in the variance of empirical loss with increasing $\tau$, which aligns with the interpretation of empirical mean-variance trade-off.

%Figure \ref{point estimation} illustrates that with larger target $\tau$, the estimated point is closer to minority points which makes the loss on samples with large loss become smaller. Figure \ref{averagevsmax} indicates that as $\tau$ increases, the average loss gradually increases, but the maximum sample loss decreases. Figure \ref{variancevstarget} indicates that as $\tau$ increases, the variance of empirical loss also decreases. The above phenomenon is consistent with what we mentioned in section \ref{sec: int}, where our KL-RS controls the spread of loss distribution to make the model more robust and improves the tail performance.

\begin{figure*}[ht] %这里使用的是强制位置，除非真的放不下，不然就是写在哪里图就放在哪里，不会乱动
  \centering  %图片全局居中
  \vspace{-0.35cm} %设置与上面正文的距离
  \subfigtopskip=2pt %设置子图与上面正文或别的内容的距离
  \subfigbottomskip=2pt %设置第二行子图与第一行子图的距离，即下面的头与上面的脚的距离
  \subfigcapskip=-5pt %设置子图与子标题之间的距离

  \includegraphics[width=0.9\linewidth]{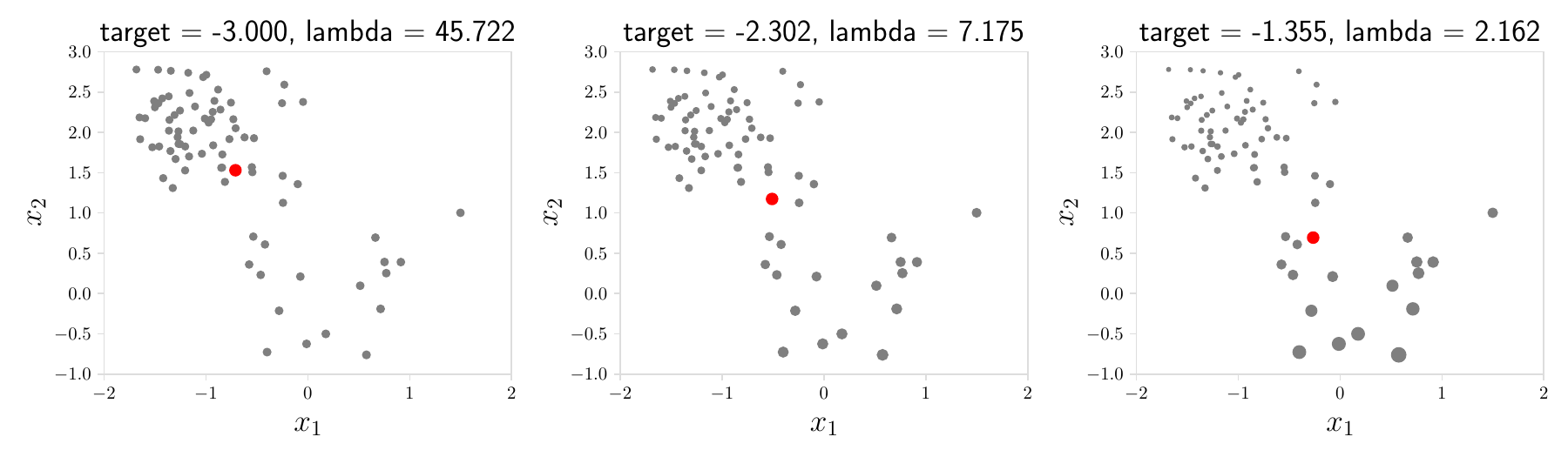}
  \caption{Weights attributed to samples}
  \label{fig4}
\end{figure*}

 We select three target $\tau$ and demonstrate the corresponding weights attributed to samples based on Proposition \ref{prop: optDistn} in Figure \ref{fig4}. The size of each data point is proportional to the weight of this sample. When $\tau=-3$, the KL-RS model assigns nearly uniform weights to all samples (left panel). 
When $\tau=-1.355$, it is obvious that the KL-RS model amplifies the influence of samples with large losses (right panel), which aligns with the interpretation of prioritizing samples with large losses.

\section{The Advantages of Feasibility Detection of the KL-RS Model} \label{sec: num kl-rs}

Our algorithm 1 demonstrates two pivotal advantages:  \textit{unbiasedness} and \textit{normalization}, which effectively avoid the bias of gradient estimation and mitigate the scaling concern.

\textbf{Problem of biasedness.} The framework of tilted empirical risk minimization (see e.g. \cite{li2023tilted, li2020tilted, qi2022attentional}) optimizes $\hat{R}(\bt, \lambda)$ directly with the gradient 
\begin{equation*}
\nabla_{\bt}\hat{R}(\bt, \lambda)=\frac{1}{\lambda} \sum_{i\in[N]}\frac{\exp(l(\bt, \hbz_{i})/\lambda)}{\sum_{j\in[N]}\exp(l(\bt, \hbz_{j})/\lambda)}\cdot \nabla_{\bt}l(\bt, \hbz_{i}).
\end{equation*}
Note that the gradient of each sample is reweighted by $\frac{1}{\lambda}\frac{\exp(l(\bt, \hbz_{i}))}{\sum_{j\in[N]}\exp(l(\bt, \hbz_{j})/\lambda)}$. The estimation of the denominator presents a significant challenge in a stochastic or mini-batch setting because a straightforward estimation of the exponential sum introduces bias. To alleviate this issue, 
a common strategy involves retaining an additional estimator for the denominator at the expense of extra sampling \cite{li2023tilted, li2020tilted, qi2022attentional}.

\textbf{Problem of scaling.}
A straightforward approach to rectifying the biased gradient estimator involves directly optimizing a component of $\hat{R}(\bt, \lambda)$:
\begin{equation*}
\min_{\bt\in\Theta} \hat{R}(\bt, \lambda) \quad \Longleftrightarrow \quad \min_{\bt\in\Theta}\mathbb{E}_{\Pemp}\left[\exp\left(\frac{l(\bt, \tbz)}{\lambda}\right)\right],
\end{equation*}
with the gradient 
\begin{equation} \label{exp: grad}
\nabla_{\bt}\mathbb{E}_{\Pemp}\left[\exp\left(\frac{l(\bt, \tbz)}{\lambda}\right)\right]=\frac{1}{\lambda N}\sum_{i\in[N]}\exp \left(\frac{l(\bt, \hbz_{i})}{\lambda} \right)\nabla_{\bt}l(\bt, \hbz_{i}).
\end{equation}
Although mini-batch gradient of \eqref{exp: grad} is an unbiased estimator, the scale of this gradient is significantly influenced by $\exp \left(\frac{l(\bt, \hbz_{i})}{\lambda} \right)$. In practice, the amplification of gradient magnitude may induce numerical errors. Conversely, a diminished gradient may cause the objective function value to decrease slowly or remain stagnant, especially when the loss is relatively small after several iterations.

\textbf{Our approach.}
The objective function of our approach is 
\begin{equation*}
\min_{\bt\in\Theta}\mathbb{E}_{\Pemp}[\exp\left(\frac{l(\bt, \tbz)-\tau}{\lambda}\right)],
\end{equation*}
with the gradient 
\begin{equation} \label{exp: our grad}
\nabla_{\bt}\mathbb{E}_{\Pemp}\left[\exp\left(\frac{l(\bt, \tbz)-\tau}{\lambda}\right)\right]=\frac{1}{\lambda N}\sum_{i\in[N]}\frac{\exp(l(\bt, \hbz_{i})/\lambda)}{\exp(\tau/\lambda)}\nabla_{\bt}l(\bt, \hbz_{i}).
\end{equation}
In a stochastic or mini-batch regime, our gradient estimator demonstrates unbiasedness. Moreover, we observe that our gradient \eqref{exp: our grad} is normalized by $\exp(\tau/\lambda)$ compared to vanilla \eqref{exp: grad}. Considering that the $t$th iterate satisfies $\sum_{i\in[N]}\exp(\frac{l(\bt_{t-1}, \hbz_{i})}{\lambda})\leq \exp(\tau/\lambda)$, the update from  $\bt_{t-1}$ to $\bt_{t}$ is more stable. 

In summary, the effective and stable performance stems from the KL-RS setting, a feature absent in the tilted empirical risk minimization framework.

\section{Algorithm development for Hierarchical KL-RS Model} \label{sec: num h kl-rs}
We consider the scope of alternative optimization. Supposing that $(\lambda_1,\lambda_2)$ is fixed, we shall verify the feasibility of $(\lambda_1,\lambda_2)$. Note that the constraint
\begin{align*}
\hat{\mathbf{R}}(\bt,\lambda_1,\lambda_2) \triangleq
&\lambda_{1}\log\left(\mathbb{E}_{\Pemp_{\tbg}}
  \left[ \exp\left(\lambda_{2}\log\left(
    \mathbb{E}_{\Pemp_{\tbz\vert\tbg}}
    \exp\left(l(\bt, \tbz)/\lambda_{2}\right)\right)/\lambda_{1}\right)\right]
  \right)    \leq \tau \\
\Longleftrightarrow \quad&   \mathbb{E}_{\Pemp_{\tbg}}
  \left[ \exp\left(\lambda_{2}\log\left(
    \mathbb{E}_{\Pemp_{\tbz\vert\tbg}}
    \exp\left(l(\bt, \tbz)/\lambda_{2}\right)\right)/\lambda_{1}\right)\right] \leq \exp(\tau/\lambda_1)\\
\Longleftrightarrow \quad& \Ex_{\Pemp_{\tbg}}\left[\exp\left(\frac{\lambda_{2}}{\lambda_{1}}\log\left(
    \Ex_{\Pemp_{\tbz\vert\tbg}}\exp(l(\bt, \tbz)/\lambda_{2})
  \right)-\frac{\tau}{\lambda_{1}}\right)\right] \leq 1\\
\Longleftrightarrow \quad& \Ex_{\Pemp_{\tbg}}\left[\exp\left(\frac{\lambda_{2}}{\lambda_{1}} \left(\log\left(
    \Ex_{\Pemp_{\tbz\vert\tbg}}\exp(l(\bt, \tbz)/\lambda_{2})
  \right)-\frac{\tau}{\lambda_{2}}\right) \right)\right] \leq 1  \\
\Longleftrightarrow \quad&    
\Ex_{\Pemp_{\tbg}}\left[\exp\left(\frac{\lambda_{2}}{\lambda_{1}} \log\left(
\Ex_{\Pemp_{\tbz\vert\tbg}}\exp\left(\frac{l(\bt, \tbz)-\tau}{\lambda_{2}}\right)
  \right) \right)\right] \leq 1 \\
\Longleftrightarrow \quad&    
\Ex_{\Pemp_{\tbg}}\left[\left(
\Ex_{\Pemp_{\tbz\vert\tbg}}\exp\left(\frac{l(\bt, \tbz)-\tau}{\lambda_{2}}\right)
  \right)^\frac{\lambda_{2}}{\lambda_{1}} \right] \leq 1.  
\end{align*}
Therefore, to assess the viability of the hierarchical KL-RS model, it is imperative to tackle the conditional stochastic optimization problem:
$$\min_{\bt} \Ex_{\Pemp_{\tbg}}\left[ h\left(
\Ex_{\Pemp_{\tbz\vert\tbg}}\left[f(\bt,\tbz;\lambda_2)\right]; \lambda_1,\lambda_2 \right) \right]$$
%\begin{equation}\label{exp: gest}
%  \nabla\hat{F}_{t}(\bt)=\frac{1}{M_{1}}\sum_{\ell\in[M_{1}]}\nabla\hat{G}_{t}(\bt, \hbg_{\ell})
%  \end{equation}
%  \begin{equation}
 % \nabla\hat{G}_{t}(\bt, \hbg_{\ell})=\nabla q_{t}(\frac{1}{M_{2}}\sum_{m\in[M_{2}]}f_{t}(\bt, \hbz_{\ell, m}))/M_{2}\sum_{m\in[M_{2}]}\nabla f_{t}(\bt, \hbz_{\ell, m})
%\end{equation}
with $h(x;\lambda_1,\lambda_2) \triangleq x^\frac{\lambda_{2}}{\lambda_{1}}$ and $f$ is defined in Section \ref{subsec: alg KL}. Generally speaking, conditional stochastic optimization problem is computationally challenging in practice, because the estimator of the gradient is often biased \cite{hu2020sample, hu2020biased, hu2021bias}. At the early stage of our experiment, we adopt the vanilla biased stochastic gradient method introduced by \cite{hu2020biased}. It appears that this method often encounters difficulties when training neural networks with batch normalization (BN) layers, especially when there are significant disparities in means and variances among different classes in Section \ref{sec: long tailed learning}. To improve and stabilize the performance, we tailor the biased  stochastic gradient method by introducing a hierarchical mini-batch strategy, resulting in the following algorithm. We can verify that the convergence rate of our algorithm aligns with the result in \cite{hu2020biased} by a constant factor.
\IncMargin{1em} % 使得行号不向外突出 
\begin{algorithm}\label{alg: feasible h-KLRS}
    \SetAlgoNoLine % 不要算法中的竖线
    \SetKwInOut{Input}{\textbf{Input}}\SetKwInOut{Output}{\textbf{Output}} % 替换关键词
    \textbf{Input:} $\lambda_1, \lambda_2$\\
    \textbf{Initialization:} group batch size $M_1$, batch size within group $M_2$, step size $\alpha$\\
\While{stopping criteria not reached}{
Sample $\hbg_{i}$ uniformly random from $\tbg$ with batch size $M_1$;\\
 \For{$i=1$ to $M_1$}{
        For each given $\hbg_{i}$, sample $\hbz_{i, j}$ uniformly random from 
        $\tbz\vert \hbg_{i}$ with batch size $M_2$;\\
        Construct  $\bar{l}_i \triangleq \frac{1}{M_2} \sum_{j \in [M_2]} l(\bt,\hbz_{i, j}), \quad \bar{\mathrm{l}}_i \triangleq \frac{1}{M_2} \sum_{j \in [M_2]} \nabla l(\bt,\hbz_{i, j})$ 
      }
    Construct $\nabla F\triangleq \frac{1}{M_1} \sum_{i\in [M_1]} h'(\bar{l}_i;\lambda_1,\lambda_2) \cdot \bar{\mathrm{l}}_i$ \\
    Update $\bt \leftarrow \bt -\alpha \nabla F$
}
\textbf{Output:} Boolean$\left(\Ex_{\Pemp_{\tbg}}\left[ h\left(
\Ex_{\Pemp_{\tbz\vert\tbg}}\left[f(\bt,\tbz;\lambda_2)\right]; \lambda_1,\lambda_2 \right) \right] \leq 1\right)$
    \caption{Feasibility of the hierarchical KL-RS model $(\lambda_1, \lambda_2)$}
\end{algorithm}
\DecMargin{1em}

Back to the scope of alternative optimization, supposing that $\bt$ is fixed, we first propose desirable properties of $\hat{\mathbf{R}}(\bt, \lambda_1,\lambda_2)$ in the following proposition.
\begin{proposition} \label{prop: h-KL func}
 $\hat{\mathbf{R}}(\bt,\lambda_1,\lambda_2)$ is convex on $(\lambda_1,\lambda_2)$, and non-increasing with respect to $\lambda_2$.
\end{proposition}
Now, supposing that $\lambda_1$ is fixed, we consider the following sub-problem \ref{prob: sub lambda 2}:
\begin{equation} \label{prob: sub lambda 2}
  \begin{aligned}
 H(\lambda_1)= &\min_{\bt,  \lambda_{2}\geq 0}\lambda_{2}\\
 \text{s.t.} \quad & \hat{\mathbf{R}}(\bt, \lambda_1,\lambda_2) \leq \tau.
  \end{aligned}
\end{equation}
Similar to Algorithm \ref{alg: bisec KL-RS}, we can develop a bisection method to solve it due to monotonicity.
\IncMargin{1em} % 使得行号不向外突出 
\begin{algorithm}\label{alg: H-bisection}
\SetAlgoNoLine % 不要算法中的竖线
\SetKwInOut{Input}{\textbf{Initialization}}\SetKwInOut{Output}{\textbf{Output}} % 替换关键词
\textbf{Input:} $\lambda_1$\\
\Input{ $\underline{\lambda}=0$, a positive value $\lambda_0$, a precision $\epsilon>0$\\}
\While{ \textnormal{Algorithm \ref{alg: feasible h-KLRS}($\lambda_1,\lambda_0$) == False}}
{$\underline{\lambda}\leftarrow \lambda_0$, $\lambda_0 \leftarrow 2\lambda_0$ }
\hspace{5mm}$\overline{\lambda}= \lambda_0$\\
\While{ $\overline{\lambda} - \underline{\lambda} \geq \epsilon$}
    {$\lambda_{\text{mid}}=(\overline{\lambda}+\underline{\lambda} )/2$\\
    \If { \textnormal{Algorithm \ref{alg: feasible h-KLRS}($\lambda_1,\lambda_{\text{mid}}$)==True} }
    {$\overline{\lambda}= \lambda_{\text{mid}}$}   \Else{$\underline{\lambda}= \lambda_{\text{mid}}$}
}
\textbf{Output: } $\lambda_{\text{mid}}$
    \caption{Find optimal $\lambda_2$ by bisection method ($\lambda_1$)}
\end{algorithm}
\DecMargin{1em}

According to the sub-problem \ref{prob: sub lambda 2}, we can reformulate the hierarchical KL-RS problem as
$$\min_{\lambda_1 \geq 0} \;\; \lambda_1+w H(\lambda_1).$$
We can verify that $\lambda_1+w H(\lambda_1)$ is a convex function on $\lambda_1$ with $w>0$ by proposition \ref{prop: h-KL func}, which can be solved by the golden-ratio search. In all, we propose our algorithm \ref{alg: H-gold} to solve the hierarchical KL-RS.
\IncMargin{1em} % 使得行号不向外突出 
\begin{algorithm}\label{alg: H-gold}
\SetAlgoNoLine % 不要算法中的竖线
\SetKwInOut{Input}{\textbf{Initialization}}\SetKwInOut{Output}{\textbf{Output}} % 替换关键词
\Input{ a precision $\epsilon>0$, $\gamma=0.382$, ${\lambda_l}=\lambda_{\min}$, ${\lambda_r}=\lambda_{\max}$\\}
\While{ $\lambda_r - \lambda_l \geq \epsilon$}
    {
    $\lambda'_l=\lambda_l+\gamma(\lambda_r-\lambda_l)$, \quad
    $\lambda'_r=\lambda_l+(1-\gamma)(\lambda_r-\lambda_l)$ \\
    $\lambda^{(2)}_l=$Algorithm \ref{alg: H-bisection}($\lambda'_l$), \quad
    $\lambda^{(2)}_r=$Algorithm \ref{alg: H-bisection}($\lambda'_r$)\\
    \If { $\lambda'_l+ w \lambda^{(2)}_l \leq \lambda'_r+ w \lambda^{(2)}_r$}
    {${\lambda_r}=\lambda'_r$}   
    \Else{${\lambda_l}=\lambda'_l$}
}
\textbf{Output: } $\lambda_{\text{mid}}$
    \caption{Solve the hierarchical KL-RS by golden-ratio search}
\end{algorithm}
\DecMargin{1em}

\section{Details for Section {\ref{sec: exp}}}
\label{app: exp}
\subsection{Details of Section \ref{sec: label distribution shift}}
\label{app: label distribution shift}
\subsubsection{Background}
Label distribution shift is a special case of the out-of-distribution (OOD) problem, where the label distribution in the training dataset does not reflect what is observed during testing. For example, this is commonly seen in predicting diseases based on symptoms. Suppose a model is trained in dermatology hospital and used in a skin disease hospital, there would be label distribution shift. Such label distribution will degrade model performance significantly. Some works assumed that we can obtain 
some unlabelled test samples so that we can use them to estimate the shift in distribution. However, unlabelled test samples are not always accessible and a trained model may be deployed on many environments. Hence, a model that is robust to label distribution shift is desired.

\subsubsection{Experiment Setting}
The HIV-1 dataset was first investigated in \cite{duchi2019variance}. 
This dataset contains 6590 samples, with 1360 positive samples and 5230 negative samples. In this experiment, we adopt a linear model with a sigmoid layer and the cross-entropy loss function. 

To obtain test datasets with varying KL divergences from the training set, we employed the following method. First, we randomly sampled 329 samples from the positive samples to construct a \textit{positive sample pool}. Then, we sampled 330 samples from the negative 
samples to construct a \textit{negative sample pool}. Then, the remaining samples were used as the training samples. Third, based on the given KL divergence value, we calculate the proportion of positive and negative samples on the test set using the distribution of 
positive and negative samples on the training set as the reference distribution. Finally, according to the calculated proportions, sample 400 instances from 
\textit{positive sample pool} and \textit{negative sample pool} to construct a test
dataset to evaluate the performance of model. Repeat the sampling process and evaluation process for multiple times with different ramdom seeds.

\textbf{Experiment platform.} The experiments are conducted on a Windows11 system with 13th Gen Intel(R) Core(TM) i7-13700H   2.40 GHz processor and 32.0 GB memory. No graphics card is used in this experiments.

\subsubsection{Hyperparameter Tuning}
We referred to \cite{li2023tilted} for selecting the hyperparameters in our benchmark. For algorithms with hyperparameters, we use the hyperparameters chosen in the original paper where the algorithm was proposed as the baseline, perform a geometric or arithmetic search over five values, and then present the results for the hyperparameter with best MCC performance. Interestingly, we initially expected that different hyperparameters might perform better at different distribution shift distances. However, the actual results often show that a particular hyperparameter tends to perform better across all distances than other hyperparameter.

For CVaRDRO, we choose 4.0 from $\{0.25, 0.50, 1.00, 2.00, 4.00\}$ as our hyperparamert. For FL, we choose 0.25 from $\{0.25, 0.5, 1.0, 2, 4\}$. For HINGE, we choose 0.5 from $\{0.5, 1, 2, 4, 8\}$. For HRM, we choose 0.00 from $\{0.00, 0.20, 0.40, 0.60, 0.80\}$. There is no hyperparameter for LR. As for our KL-RS, we set $\tau=(1+\epsilon)E_0$ with $\{0.10, 0.20, 0.30, 0.40, 0.50\}$. We present the result with $\epsilon=0.10$ and $\epsilon=0.50$.

\subsubsection{Computation Time}
For each algorithm, we update 10000 times. We conduct each optimization algorithm five times and report the mean and standard deviation of the computation time in table
\ref{tab: computation time LDS}. We do not present the computation time for HINGE because it is computed using the `LinearSVC` package from `sklearn.svm`. The other algorithms are based on gradient descent, making the computation time for HINGE not directly comparable.

Even thought the computation time of KL-RS is longer than ERM, compared with other benchmarks, extra computation burdern is mild.

\begin{table}[H]
\centering
\caption{Computation Time for Label Distribution Shift}
\label{tab: computation time LDS}
\begin{tabular}{cccccccc} % 五列对齐方式
    \toprule
    Algorithm & KL-RS0.10 & KL-RS0.50 & ERM & CVaRDRO & FL & HRM & LR \\
    \hline
    Average Time(s) & 5.69 & 5.62 & 2.66 & 16.186 & 5.88 & 21.66 & 49.17\\
    \hline
    Std Time(s) & 0.11 & 0.19 & 0.06 & 0.22 & 0.41 & 0.10& 0.41 \\
    \bottomrule
\end{tabular}
\end{table}

\subsubsection{Evaluation Metric}
We adopt a variety of metrics to evaluate the model's performance, including F1 score, Matthews Correlation Coefficient (MCC), accuracy on rare samples (Acc1),
accuracy on common sample (Acc0), and accuracy on overall samples (Acc). Particularly, MCC is an evaluation metric tailored for binary classification task with imbalanced 
labels with the following definition, 
\begin{definition}[Matthews Correlation Coefficient (MCC)]
  \begin{equation}
    MCC\triangleq \frac{TP\times TN - FP\times FN}{\sqrt{(TP+FP)
    (TP+FN)(TN+FP)(TN+FN)}},
  \end{equation}
where $TP$ denotes true positive samples, $TN$ denotes true negative samples, $FP$ denotes false positive samples, and $FN$ denotes false negative samples.
\end{definition}

In addition to the classic binary classification evaluation metrics
mentioned earlier, we also employ rank error as a metric.

\begin{definition}[Rank Error]

    Given a prediction model $h$, a positive sample $x_{+}$ and a negative sample 
    $x_{-}$, the ranking error is defined as 
    \begin{equation}
        \epsilon(h)\triangleq h(x_{-})-h(x_{+}).
    \end{equation}
\end{definition}
For an ideal classifier, it should always assign a higher score to a positive sample 
compared to a negative sample. When model assigns a larger value to a negative sample,
the model makes some mistakes. Smaller rank error is preferred by a binary 
classification model. 

We investigate the Value-at-Risk (VaR) of rank 
error  which is the $\alpha$-quantile of rank error. For a r.v. 
$\tilde{v}$, its $\alpha$-quantile is $VaR_\alpha(\tilde{v})\triangleq\min
\{a\vert\mathbb{P}(\tilde{v}\leq a)\geq \alpha\}$ and its $\alpha$-superquantile 
is $CVaR_\alpha(\tilde{v})\triangleq\mathbb{E}[\tbz\vert \tbz\geq 
VaR_\alpha(\tilde{v} )]$. Both $VaR_\alpha(\epsilon(h))$ and 
$CVaR_\alpha(\epsilon(h))$ evaluate the risk of large rank error, 
which can be interpreted as risk of misclassification. We adopt 
$VaR_{90}(\epsilon(h))$ and $CVaR_{90}(\epsilon(h))$ as our metric 
to evaluate our model. 

We evaluate the model's performance on test sets at 21 different KL distances 
$\{0.00, 0.01, \cdots, 0.20\}$.
For each given distance, we employ 5 random seeds to randomly sample 
the test sets and record the average metrics. We present the experimental 
results at distance of 0.00, 0.05, 0.10, 0.15 and 0.20 in Table \ref{tab: kl000}, \ref{tab: kl005}, \ref{tab: kl010}, \ref{tab: kl015} and \ref{tab: kl020}.
We select results from these distances to demonstrate outcomes in three scenarios: 
no distribution shift, slight shift, and significant shift.

As we have mentioned, KL-RS sacrifices some performance on average to improve
the model's generalization capability. 

As the KL distance increases, various metrics for KL-RS0.10 degrade significantly, similar to ERM. However, KL-RS0.10 completely outperforms the ERM approach by achieving comparable in-sample Acc, MCC, and F1 scores, while also outperforming ERM on all metrics except for Acc0 after the distribution shift. KL-RS0.10 achieves a noticeable improvement in generalization with a negligible loss in in-sample performance.

Compared with KLRS0.10, KLRS0.50 improves the generalization ability of model greatly. The KL-RS0.50 model's performance does not deteriorate rapidly with 
distribution shifts. It is foreseeable that, as the distance further increases, the various metrics for KL-RS0.50 will not experience significant degradation. However, this strong generalization ability comes at the cost of reduced in-sample performance. When there is no distribution shift or only a slight shift, KL-RS0.50 underperforms compared to ERM in terms of MCC, F1, and Acc.

\begin{table*}[ht]
\centering
\caption{Results Of Label Distribution Shift at Distance 0.00}
\label{tab: kl000}
\begin{tabular}{|c|c|c|c|c|c|c|c|c|}
  \hline
  \multicolumn{1}{|c|}{\multirow{2}{*}{Algorithm}} &\multicolumn{1}{c|}{\multirow{2}{*}{Statistic}}& \multicolumn{7}{c|}{Metric}\\
  \cline{3-9}
   & &\multicolumn{1}{c|}{Acc1}&\multicolumn{1}{c|}{Acc0}&\multicolumn{1}{c|}{Acc}&\multicolumn{1}{c|}{MCC}&
  \multicolumn{1}{|c|}{F1}&\multicolumn{1}{c|}{$\mbox{VaR}_{90}$}&\multicolumn{1}{c|}{$\mbox{CVaR}_{90}$}\\
  \hline
   \multicolumn{1}{|c|}{
   \multirow{2}{*}{ERM} } & mean & 0.807 & 0.973 & 0.917 & 0.814 & 0.868 & -0.189 & -0.028  \\
  \cline{2-2}
   & std & $\pm0.026$ & $\pm0.003$ & $\pm0.009$ & $\pm0.020$ & $\pm0.016$ & $\pm0.040$ & $\pm0.027$\\
  \hline
  \multicolumn{1}{|c|}{
   \multirow{2}{*}{CVaRDRO} } & mean & 0.875 & 0.890 & 0.885 & 0.751 & 0.838 & -0.009 & -0.004 \\
  \cline{2-2}
   & std & $\pm0.018$ & $\pm0.006$ & $\pm0.006$ & $\pm0.013$ & $\pm0.009$ & $\pm0.001$ & $\pm0.001$\\
  \hline
  \multicolumn{1}{|c|}{
   \multirow{2}{*}{KL-RS0.10} }& mean & 0.832 & 0.960 & 0.916 & 0.812 & 0.871 & -0.232 & -0.048 \\
  \cline{2-2}
   & std & $\pm0.018$ & $\pm0.002$ & $\pm0.007$ & $\pm0.015$ & $\pm0.011$ & $\pm0.044$ & $\pm0.031$\\
  \hline
  \multicolumn{1}{|c|}{
   \multirow{2}{*}{KL-RS0.50} } & mean & 0.887 & 0.888 & 0.887 & 0.758 & 0.843 & -0.354 & -0.133 \\
  \cline{2-2}
   & std & $\pm0.014$ & $\pm0.007$ & $\pm0.004$ & $\pm0.009$ & $\pm0.006$ & $\pm0.038$ & $\pm0.040$\\
  \hline
  \multicolumn{1}{|c|}{
   \multirow{2}{*}{HRM} } & mean & 0.812 & 0.970 & 0.916 & 0.812 & 0.868 & -0.188 & -0.027 \\
  \cline{2-2}
   & std & $\pm0.024$ & $\pm0.004$ & $\pm0.009$ & $\pm0.020$ & $\pm0.015$ & $\pm0.041$ & $\pm0.027$\\
  \hline
  \multicolumn{1}{|c|}{
   \multirow{2}{*}{FL} } & mean & 0.812 & 0.973 & 0.918 & 0.817 & 0.871 & -0.200 & -0.043 \\
  \cline{2-2}
   & std & $\pm0.024$ & $\pm0.003$ & $\pm0.008$ & $\pm0.018$ & $\pm0.014$ & $\pm0.036$ & $\pm0.027$\\
  \hline
  \multicolumn{1}{|c|}{
   \multirow{2}{*}{HINGE} } & mean & 0.825 & 0.962 & 0.915 & 0.810 & 0.869 & -0.151 & -0.030 \\
  \cline{2-2}
   & std & $\pm0.023$ & $\pm0.003$ & $\pm0.008$ & $\pm0.018$ & $\pm0.014$ & $\pm0.015$ & $\pm0.017$\\
  \hline
  \multicolumn{1}{|c|}{
   \multirow{2}{*}{LR} } & mean & 0.834 & 0.964 & 0.920 & 0.820 & 0.876 & -0.232 & -0.048 \\
   \cline{2-2}
    & std & $\pm0.021$ & $\pm0.003$ & $\pm0.009$ & $\pm0.020$ & $\pm0.015$ & $\pm0.044$ & $\pm0.032$\\
  \hline
\end{tabular}
\end{table*}

\begin{table*}[ht]
\centering
\caption{Results Of Label Distribution Shift at Distance 0.05}
\label{tab: kl005}
\begin{tabular}{|c|c|c|c|c|c|c|c|c|}
  \hline
  \multicolumn{1}{|c|}{\multirow{2}{*}{Algorithm}} &\multicolumn{1}{c|}{\multirow{2}{*}{Statistic}}& \multicolumn{7}{c|}{Metric}\\
  \cline{3-9}
   & &\multicolumn{1}{c|}{Acc1}&\multicolumn{1}{c|}{Acc0}&\multicolumn{1}{c|}{Acc}&\multicolumn{1}{c|}{MCC}&
  \multicolumn{1}{|c|}{F1}&\multicolumn{1}{c|}{$\mbox{VaR}_{90}$}&\multicolumn{1}{c|}{$\mbox{CVaR}_{90}$}\\
  \hline
   \multicolumn{1}{|c|}{
   \multirow{2}{*}{ERM} } & mean & 0.806 & 0.972 & 0.882 & 0.780 & 0.880 & -0.169 & -0.021 \\
  \cline{2-2}
   & std & $\pm0.019$ & $\pm0.006$ & $\pm0.010$ & $\pm0.017$ & $\pm0.011$ & $\pm0.019$ & $\pm0.015$\\
  \hline
  \multicolumn{1}{|c|}{
   \multirow{2}{*}{CVaRDRO} } & mean & 0.880 & 0.899 & 0.889 & 0.778 & 0.895 & -0.008 & -0.003 \\
  \cline{2-2}
   & std & $\pm0.008$ & $\pm0.016$ & $\pm0.004$ & $\pm0.009$ & $\pm0.003$ & $\pm0.001$ & $\pm0.001$\\
  \hline
  \multicolumn{1}{|c|}{
   \multirow{2}{*}{KL-RS0.10} }& mean & 0.832 & 0.961 & 0.891 & 0.793 & 0.892 & -0.214 & -0.041 \\
  \cline{2-2}
   & std & $\pm0.017$ & $\pm0.007$ & $\pm0.009$ & $\pm0.016$ & $\pm0.010$ & $\pm0.022$ & $\pm0.018$\\
  \hline
  \multicolumn{1}{|c|}{
   \multirow{2}{*}{KL-RS0.50} } & mean & 0.894 & 0.899 & 0.896 & 0.793 & 0.903 & -0.357 & -0.133 \\
  \cline{2-2}
   & std & $\pm0.011$ & $\pm0.019$ & $\pm0.007$ & $\pm0.014$ & $\pm0.006$ & $\pm0.027$ & $\pm0.021$\\
  \hline
  \multicolumn{1}{|c|}{
   \multirow{2}{*}{HRM} } & mean & 0.806 & 0.970 & 0.881 & 0.778 & 0.880 & -0.168 & -0.020 \\
  \cline{2-2}
   & std & $\pm0.016$ & $\pm0.006$ & $\pm0.008$ & $\pm0.013$ & $\pm0.009$ & $\pm0.019$ & $\pm0.015$\\
  \hline
  \multicolumn{1}{|c|}{
   \multirow{2}{*}{FL} } & mean & 0.810 & 0.972 & 0.885 & 0.784 & 0.883 & -0.186 & -0.037 \\
  \cline{2-2}
   & std & $\pm0.016$ & $\pm0.006$ & $\pm0.008$ & $\pm0.015$ & $\pm0.010$ & $\pm0.017$ & $\pm0.016$\\
  \hline
  \multicolumn{1}{|c|}{
   \multirow{2}{*}{HINGE} } & mean & 0.821 & 0.964 & 0.887 & 0.786 & 0.887 & -0.152 & -0.031 \\
  \cline{2-2}
   & std & $\pm0.013$ & $\pm0.010$ & $\pm0.006$ & $\pm0.011$ & $\pm0.006$ & $\pm0.017$ & $\pm0.014$\\
  \hline
  \multicolumn{1}{|c|}{
   \multirow{2}{*}{LR} } & mean & 0.836 & 0.968 & 0.897 & 0.804 & 0.897 & -0.217 & -0.041 \\
   \cline{2-2}
    & std & $\pm0.015$ & $\pm0.004$ & $\pm0.008$ & $\pm0.013$ & $\pm0.008$ & $\pm0.026$ & $\pm0.019$\\
  \hline
\end{tabular}
\end{table*}

\begin{table*}[ht]
\centering
\caption{Results Of Label Distribution Shift at Distance 0.10}
\label{tab: kl010}
\begin{tabular}{|c|c|c|c|c|c|c|c|c|}
  \hline
  \multicolumn{1}{|c|}{\multirow{2}{*}{Algorithm}} &\multicolumn{1}{c|}{\multirow{2}{*}{Statistic}}& \multicolumn{7}{c|}{Metric}\\
  \cline{3-9}
   & &\multicolumn{1}{c|}{Acc1}&\multicolumn{1}{c|}{Acc0}&\multicolumn{1}{c|}{Acc}&\multicolumn{1}{c|}{MCC}&
  \multicolumn{1}{|c|}{F1}&\multicolumn{1}{c|}{$\mbox{VaR}_{90}$}&\multicolumn{1}{c|}{$\mbox{CVaR}_{90}$}\\
  \hline
   \multicolumn{1}{|c|}{
   \multirow{2}{*}{ERM} } & mean & 0.798 & 0.970 & 0.863 & 0.744 & 0.878 & -0.164 & -0.009 \\
  \cline{2-2}
   & std & $\pm0.018$ & $\pm0.010$ & $\pm0.012$ & $\pm0.020$ & $\pm0.012$ & $\pm0.024$ & $\pm0.026$\\
  \hline
  \multicolumn{1}{|c|}{
   \multirow{2}{*}{CVaRDRO} } & mean & 0.881 & 0.899 & 0.888 & 0.768 & 0.907 & -0.008 & -0.003 \\
  \cline{2-2}
   & std & $\pm0.007$ & $\pm0.017$ & $\pm0.008$ & $\pm0.018$ & $\pm0.007$ & $\pm0.001$ & $\pm0.001$\\
  \hline
  \multicolumn{1}{|c|}{
   \multirow{2}{*}{KL-RS0.10} }& mean & 0.822 & 0.951 & 0.871 & 0.751 & 0.888 & -0.206 & -0.027 \\
  \cline{2-2}
   & std & $\pm0.016$ & $\pm0.018$ & $\pm0.014$ & $\pm0.028$ & $\pm0.013$ & $\pm0.031$ & $\pm0.031$\\
  \hline
  \multicolumn{1}{|c|}{
   \multirow{2}{*}{KL-RS0.50} } & mean & 0.892 & 0.894 & 0.893 & 0.777 & 0.912 & -0.323 & -0.106 \\
  \cline{2-2}
   & std & $\pm0.009$ & $\pm0.022$ & $\pm0.012$ & $\pm0.026$ & $\pm0.010$ & $\pm0.029$ & $\pm0.032$\\
  \hline
  \multicolumn{1}{|c|}{
   \multirow{2}{*}{HRM} } & mean & 0.798 & 0.967 & 0.862 & 0.741 & 0.878 & -0.163 & -0.008 \\
  \cline{2-2}
   & std & $\pm0.015$ & $\pm0.010$ & $\pm0.011$ & $\pm0.020$ & $\pm0.011$ & $\pm0.024$ & $\pm0.026$\\
  \hline
  \multicolumn{1}{|c|}{
   \multirow{2}{*}{FL} } & mean & 0.802 & 0.970 & 0.865 & 0.748 & 0.881 & -0.180 & -0.024 \\
  \cline{2-2}
   & std & $\pm0.015$ & $\pm0.010$ & $\pm0.010$ & $\pm0.018$ & $\pm0.010$ & $\pm0.022$ & $\pm0.026$\\
  \hline
  \multicolumn{1}{|c|}{
   \multirow{2}{*}{HINGE} } & mean & 0.816 & 0.955 & 0.868 & 0.749 & 0.885 & -0.139 & -0.021 \\
  \cline{2-2}
   & std & $\pm0.013$ & $\pm0.012$ & $\pm0.011$ & $\pm0.021$ & $\pm0.010$ & $\pm0.030$ & $\pm0.021$\\
  \hline
  \multicolumn{1}{|c|}{
   \multirow{2}{*}{LR} } & mean & 0.828 & 0.959 & 0.878 & 0.765 & 0.894 & -0.205 & -0.026 \\
   \cline{2-2}
    & std & $\pm0.016$ & $\pm0.016$ & $\pm0.013$ & $\pm0.026$ & $\pm0.012$ & $\pm0.033$ & $\pm0.029$\\
  \hline
\end{tabular}
\end{table*}

\begin{table*}[ht]
\centering
\caption{Results Of Label Distribution Shift at Distance 0.15}
\label{tab: kl015}
\begin{tabular}{|c|c|c|c|c|c|c|c|c|}
  \hline
  \multicolumn{1}{|c|}{\multirow{2}{*}{Algorithm}} &\multicolumn{1}{c|}{\multirow{2}{*}{Statistic}}& \multicolumn{7}{c|}{Metric}\\
  \cline{3-9}
   & &\multicolumn{1}{c|}{Acc1}&\multicolumn{1}{c|}{Acc0}&\multicolumn{1}{c|}{Acc}&\multicolumn{1}{c|}{MCC}&
  \multicolumn{1}{|c|}{F1}&\multicolumn{1}{c|}{$\mbox{VaR}_{90}$}&\multicolumn{1}{c|}{$\mbox{CVaR}_{90}$}\\
  \hline
   \multicolumn{1}{|c|}{
   \multirow{2}{*}{ERM} } & mean & 0.801 & 0.966 & 0.853 & 0.716 & 0.882 & -0.162 & -0.008 \\
  \cline{2-2}
   & std & $\pm0.013$ & $\pm0.007$ & $\pm0.010$ & $\pm0.017$ & $\pm0.009$ & $\pm0.022$ & $\pm0.026$\\
  \hline
  \multicolumn{1}{|c|}{
   \multirow{2}{*}{CVaRDRO} } & mean & 0.886 & 0.894 & 0.888 & 0.753 & 0.916 & -0.008 & -0.003 \\
  \cline{2-2}
   & std & $\pm0.007$ & $\pm0.045$ & $\pm0.016$ & $\pm0.041$ & $\pm0.011$ & $\pm0.001$ & $\pm0.001$\\
  \hline
  \multicolumn{1}{|c|}{
   \multirow{2}{*}{KL-RS0.10} }& mean & 0.830 & 0.955 & 0.869 & 0.737 & 0.897 & -0.203 & -0.024 \\
  \cline{2-2}
   & std & $\pm0.010$ & $\pm0.011$ & $\pm0.008$ & $\pm0.015$ & $\pm0.007$ & $\pm0.028$ & $\pm0.031$\\
  \hline
  \multicolumn{1}{|c|}{
   \multirow{2}{*}{KL-RS0.50} } & mean & 0.897 & 0.892 & 0.895 & 0.766 & 0.922 & -0.319 & -0.097 \\
  \cline{2-2}
   & std & $\pm0.009$ & $\pm0.037$ & $\pm0.014$ & $\pm0.034$ & $\pm0.010$ & $\pm0.032$ & $\pm0.027$\\
  \hline
  \multicolumn{1}{|c|}{
   \multirow{2}{*}{HRM} } & mean & 0.801 & 0.966 & 0.853 & 0.716 & 0.882 & -0.162 & -0.006 \\
  \cline{2-2}
   & std & $\pm0.012$ & $\pm0.007$ & $\pm0.009$ & $\pm0.016$ & $\pm0.008$ & $\pm0.023$ & $\pm0.026$\\
  \hline
  \multicolumn{1}{|c|}{
   \multirow{2}{*}{FL} } & mean & 0.807 & 0.966 & 0.856 & 0.721 & 0.885 & -0.178 & -0.022 \\
  \cline{2-2}
   & std & $\pm0.010$ & $\pm0.007$ & $\pm0.008$ & $\pm0.014$ & $\pm0.007$ & $\pm0.020$ & $\pm0.027$\\
  \hline
  \multicolumn{1}{|c|}{
   \multirow{2}{*}{HINGE} } & mean & 0.820 & 0.958 & 0.863 & 0.729 & 0.892 & -0.139 & -0.019 \\
  \cline{2-2}
   & std & $\pm0.009$ & $\pm0.009$ & $\pm0.008$ & $\pm0.014$ & $\pm0.006$ & $\pm0.030$ & $\pm0.018$\\
  \hline
  \multicolumn{1}{|c|}{
   \multirow{2}{*}{LR} } & mean & 0.833 & 0.960 & 0.872 & 0.745 & 0.900 & -0.201 & -0.024 \\
   \cline{2-2}
    & std & $\pm0.010$ & $\pm0.008$ & $\pm0.007$ & $\pm0.012$ & $\pm0.006$ & $\pm0.030$ & $\pm0.030$\\
  \hline
\end{tabular}
\end{table*}

\begin{table*}[ht]
\centering
\caption{Results Of Label Distribution Shift at Distance 0.20}
\label{tab: kl020}
\begin{tabular}{|c|c|c|c|c|c|c|c|c|}
  \hline
  \multicolumn{1}{|c|}{\multirow{2}{*}{Algorithm}} &\multicolumn{1}{c|}{\multirow{2}{*}{Statistic}}& \multicolumn{7}{c|}{Metric}\\
  \cline{3-9}
   & &\multicolumn{1}{c|}{Acc1}&\multicolumn{1}{c|}{Acc0}&\multicolumn{1}{c|}{Acc}&\multicolumn{1}{c|}{MCC}&
  \multicolumn{1}{|c|}{F1}&\multicolumn{1}{c|}{$\mbox{VaR}_{90}$}&\multicolumn{1}{c|}{$\mbox{CVaR}_{90}$}\\
  \hline
   \multicolumn{1}{|c|}{
   \multirow{2}{*}{ERM} } & mean & 0.803 & 0.964 & 0.844 & 0.683 & 0.885 & -0.159 & 0.000 \\
  \cline{2-2}
   & std & $\pm0.007$ & $\pm0.019$ & $\pm0.010$ & $\pm0.022$ & $\pm0.007$ & $\pm0.022$ & $\pm0.028$\\
  \hline
  \multicolumn{1}{|c|}{
   \multirow{2}{*}{CVaRDRO} } & mean & 0.883 & 0.917 & 0.891 & 0.746 & 0.924 & -0.008 & -0.003 \\
  \cline{2-2}
   & std & $\pm0.006$ & $\pm0.021$ & $\pm0.008$ & $\pm0.020$ & $\pm0.006$ & $\pm0.001$ & $\pm0.001$\\
  \hline
  \multicolumn{1}{|c|}{
   \multirow{2}{*}{KL-RS0.10} }& mean & 0.831 & 0.950 & 0.861 & 0.704 & 0.899 & -0.199 & -0.016 \\
  \cline{2-2}
   & std & $\pm0.005$ & $\pm0.020$ & $\pm0.008$ & $\pm0.020$ & $\pm0.006$ & $\pm0.028$ & $\pm0.031$\\
  \hline
  \multicolumn{1}{|c|}{
   \multirow{2}{*}{KL-RS0.50} } & mean & 0.895 & 0.913 & 0.900 & 0.760 & 0.930 & -0.311 & -0.085 \\
  \cline{2-2}
   & std & $\pm0.005$ & $\pm0.019$ & $\pm0.006$ & $\pm0.017$ & $\pm0.004$ & $\pm0.021$ & $\pm0.035$\\
  \hline
  \multicolumn{1}{|c|}{
   \multirow{2}{*}{HRM} } & mean & 0.802 & 0.960 & 0.842 & 0.679 & 0.884 & -0.158 & 0.001 \\
  \cline{2-2}
   & std & $\pm0.006$ & $\pm0.019$ & $\pm0.009$ & $\pm0.021$ & $\pm0.006$ & $\pm0.022$ & $\pm0.028$\\
  \hline
  \multicolumn{1}{|c|}{
   \multirow{2}{*}{FL} } & mean & 0.808 & 0.964 & 0.847 & 0.689 & 0.888 & -0.176 & -0.015 \\
  \cline{2-2}
   & std & $\pm0.006$ & $\pm0.019$ & $\pm0.008$ & $\pm0.020$ & $\pm0.006$ & $\pm0.020$ & $\pm0.027$\\
  \hline
  \multicolumn{1}{|c|}{
   \multirow{2}{*}{HINGE} } & mean & 0.821 & 0.952 & 0.854 & 0.694 & 0.894 & -0.135 & -0.015 \\
  \cline{2-2}
   & std & $\pm0.006$ & $\pm0.013$ & $\pm0.007$ & $\pm0.016$ & $\pm0.005$ & $\pm0.029$ & $\pm0.016$\\
  \hline
  \multicolumn{1}{|c|}{
   \multirow{2}{*}{LR} } & mean & 0.834 & 0.956 & 0.865 & 0.713 & 0.902 & -0.196 & -0.015 \\
   \cline{2-2}
    & std & $\pm0.005$ & $\pm0.015$ & $\pm0.007$ & $\pm0.017$ & $\pm0.005$ & $\pm0.028$ & $\pm0.030$\\
  \hline
\end{tabular}
\end{table*}

\subsection{Details of section \ref{sec: long tailed learning}}
\label{app: long tailed learning}
\subsubsection{Background}
Long-Tailed Learning (LTL) is a challenge faced by machine learning when applied to real-word dataset, referring to a situation where a minority of classes dominate the training data with a large number of samples, while majority of classes are underrepresented
with only a few samples \cite{kang2020exploring, liu2019large, 
cui2019class}. A model trained on training data with a long-tailed class distribution can be easily biased towards dominant classes and suffer high loss on tail classes. To address this issue, some class sensitive learning algorithm is proposed \cite{ren2018learning, zhou2005training,
zhao2018adaptive, cui2019class}. The Label-Distribution-Aware-Margin (LDAM) technique is a re-margining approach that introduces class-dependent margin factors for distinct classes, determined by their respective frequencies in the training labels \cite{cao2019learning}. This strategy incentivizes classes with lower frequencies to possess larger margins. LDAM achieves state-of-the-art in long tailed learning as reported in \cite{zhang2023deep}.
Focal loss represents a re-weighting approach that leverages prediction probabilities to inversely adjust the weights assigned to classes\cite{lin2017focal}. Consequently, it allocates higher weights to the more challenging tail classes while assigning lower weights to the comparatively easier head classes. These algorithms often requires a prior knowledge of the size of each class.

\subsubsection{Formulation}
We introduce an extra problem formulation Group KL-RS. It is an special case of our Hierarchical KL-RS \eqref{eq: hierachicalKL-RS}. In practical applications, people often do not simultaneously address 
shifts in group and individual. Many works assume that there is no 
distribution shift happens to $\tbz\vert\tbg$ 
\cite{sagawa2019distributionally, zhang2021coping, arjovsky2019invariant, 
duchi2023distributionally}. Under this assumption we have $D_{KL}(\mathbb{P}_{\tbz\vert\tbg}\Vert\Pemp_{\tbz\vert\tbg})=0$ and the optimal value of
\eqref{eq: hierachicalKL-RS} is achieved when $\lambda_{2}=0$.
In this case, our \eqref{eq: hierachicalKL-RS} will degenerate into 
Group KL-RS formulation defined as following:
  \begin{equation}\label{eq: groupKL-RS}
    \begin{aligned}
      &\min_{\bt\in\Theta, \lambda\geq 0}\lambda\\
      &\text{s.t.}\lambda\log\left(
        \mathbb{E}_{\Pemp_{\tbg}}\exp\left(
        \mathbb{E}_{\Pemp_{\tbz\vert \tbg}}
        [l(\bt, \tbz)/\lambda]\right)\right)\leq \tau.
    \end{aligned}
  \end{equation}
We can apply our Group KL-RS \eqref{eq: groupKL-RS} to address this problem. Suppose $\tbz = (\tbx, \ty)$ is the joint r.v. of feature r.v. $\tbx$ and label r.v.
$\ty$. As we discussed in
section \ref{sec: long tailed learning} distribution shift only happens
at class proportion level. We can set $\tilde{g}=\ty$ and our Group KL-RS
\eqref{eq: groupKL-RS} has the following formulation:

Our Group KL-RS can also address this issue and does not need the prior information.
Suppose $\tbz = (\tbx, \ty)$ is the joint r.v. of feature r.v. $\tbx$ and label r.v.
$\ty$. Consider our Group KL-RS formulation \eqref{eq: groupKL-RS} and let $\tilde{g}=\ty$, 
then we can obtain the following formulation:

\begin{equation}\label{Prob: labeleq}
  \begin{aligned}
      &\min_{\bt\in\Theta, \lambda\geq 0}\lambda\\
      &\text{s.t.}\lambda\log\left(\mathbb{E}_{\Pemp_{\ty}}
      \exp\left(\mathbb{E}_{\Pemp_{\tbx\vert\ty}}
      l(\bt, (\tbx, \ty))
      /\lambda\right)\right)\leq \tau
  \end{aligned}
\end{equation}
\subsubsection{Experiment Setting}

In ERM, KLRS, CVaRDRO experiments, we use cross entropy loss as our loss function.
In Ldam and Focal experiments, we use corresponding loss functions. 
In KLRSLdam and KLRSFocal, we adopt Ldam loss and focal loss respectively and use our 
Group KL-RS paradigm to further improve model performance.

\textbf{Experiment platform.} The experiment platform is a Ubuntu Server 18.04.01 with 32G RAM an Intel(R) Xeon(R) W-2140B CPU @ 3,20 GHz, which has 8 cores and 16 threads. The GPU is GeForce GTX 1080 Ti and CuDA version is 11.4. 

\subsubsection{Computation Time}

We report the computation time in CIFAR10(LT). For each algorithm, we implement the algorithm with 200 epochs five times and report the mean and std of computation time. An interesting observation is that, in this experiment, KL-RS does not have a significantly longer computation time compared to ERM. However in table \ref{tab: computation time LDS}, the computation time of KL-RS is much longer than ERM. One possible reason is that the experiment was implemented using the PyTorch framework, which includes optimizations to reduce sampling and computation time. It is possible that the time required for KL-RS's bisection process overlapped with other steps, thereby minimizing its overall impact on computation time.

\begin{table}[H]
\centering
\caption{Computation Time for Label Distribution Shift CIFAR-10(LT) with $\rho=0.1$}
\label{tab: computation time 01}
\begin{tabular}{ccccccc} % 五列对齐方式
    \toprule
    Algorithm & ERM & KL-RS & Focal & KL-RS Focal & Ldam & KL-RS Ldam \\
    \hline
    Average Time(s) & 915.80 & 914.06 & 910.29 & 913.08 & 911.53 & 918.97 \\
    \hline
    Std Time(s) & 14.70 & 14.85 & 15.30 & 12.55 & 16.38 & 10.44 \\
    \bottomrule
\end{tabular}
\end{table}

\begin{table}[H]
\centering
\caption{Computation Time for Label Distribution Shift CIFAR-10(LT) with $\rho=0.01$}
\label{tab: computation time 001}
\begin{tabular}{ccccccc} % 五列对齐方式
    \toprule
    Algorithm & ERM & KL-RS & Focal & KL-RS Focal & Ldam & KL-RS Ldam \\
    \hline
    Average Time(s) & 601.93 & 599.19 & 598.22 & 601.51 & 601.94 & 604.23 \\
    \hline
    Std Time(s) & 7.95 & 3.47 & 5.94 & 5.79 & 5.29 & 3.18 \\
    \bottomrule
\end{tabular}
\end{table}

\subsubsection{Other Discussion About The Result}

In the main text, we mentioned that KL-RS can enhance model performance and 
further improve performance when combined with other methods. 
In addition to the experimental results, there are other 
phenomena that we analyze here.

It is worth noting that experimental results reported at Table \ref{tab: lll} indicate that KL-RS does not necessarily improve worst-case accuracy and may even perform worse than ERM. This is due to characteristics of the data itself. Our formulation \eqref{Prob: labeleq} assumes that the conditional distribution $\tilde{x}\vert\tilde{y}$ of the test
dataset is similar to the one of train dataset. However, in reality,
on the CIFAR dataset, there exist a shift on conditional distribution $\tilde{x}\vert\tilde{y}$. 
When the labels are evenly distributed in the training set, there are some 
classes where the model achieves high accuracy on the training set but performs 
very poorly on the test set. This phenomenon occurs precisely because there is 
a shift in the conditional distribution. Some correlations between labels and 
features that exist in the training set may not necessarily exist in the 
test set.

When we adopt \eqref{Prob: labeleq} to handle the long tail distribution trainset, the worst 
performance class is usually the class with the fewest samples. Problem \eqref{Prob: labeleq} pays more attention to this class. However, the shift in conditional probabilities may cause the class with the worst performance on the test set to differ from the class with the worst performance on the training set. The worst performance class on test dataset may be the one with significant distribution shift instead of the rare class. In other words, \eqref{Prob: labeleq} focus on a wrong class. While this phenomenon is indeed worth studying, such issues are beyond the scope of this paper's discussion.

Another phenomenon is observed in CIFAR-100 (LT) when the imbalance factor $\rho$ is set to 0.01. In this setting, the worst class accuracy is all $0.00$. The underlying reason behind this phenomenon may be that, under this setting, the class with the fewest samples in the training set contains only five samples. The extremely limited number of samples may require alternative handling methods.

\subsection{Details of section \ref{sec: fair pca}}

\label{app: fair pca}

\subsubsection{Background}

Decisions produced by machine learning algorithms are often skewed toward a particular group which is due to either 'biased data' or 'biased algorithm' bins \cite{barocas2016big, mehrabi2021survey, pessach2022review}. One popular approach to addressing this issue is optimizing for the worst-case
scenarios. The philosophy behind this criterion is Rawls's \textit{difference
principle}, where optimizing the worst-off group is fair since it ensures the 
minorities consent to\cite{rawls2001justice,hashimoto2018fairness,samadi2018price, 
tantipongpipat2019multi}.
The original mathematical formulation for Fair PCA in \cite{samadi2018price}
only considers the situation where there only exist two subgroups. This definition 
is later extended in \cite{tantipongpipat2019multi} to encompass scenarios involving multiple subgroups.

Recall that we adopt the reconstruction error as our loss function in this section. For a matrix $Y\in\mathbb{R}^{a\times n}$, the reconstruct error by projecting it to a rank-$d$ matrix $Z\in\mathbb{R}^{a\times n}$ is given by
\begin{equation*}
loss(Y, Z)\triangleq\Vert Y-Z\Vert_{F}^{2}-\Vert Y-\hat{Y}\Vert_{F}^{2},
\end{equation*}
where $\hat{Y}\in\mathbb{R}^{a\times n}$ is the optimal rank-d 
approximation of $Y$.

\begin{definition}[Fair PCA]

    Given m data points in $\mathbb{R}^{n}$ with  $J$ subgroups $\{A_{j}\}_{j\in[J]}$, 
    we define the problem of a fair PCA projection into d-dimensions as
    \begin{equation*}
      \min_{U\in\mathbb{R}^{m\times n}, rank(U)\leq d} 
      \max_{j\in[J]}\left\{\frac{1}{\vert A_{j}\vert} loss(A_{j}, A_{j}UU^{T})\right\}.
    \end{equation*}
\end{definition}

\begin{definition}[KL-RS PCA]
    Given m data points in $\mathbb{R}^{n}$ with $J$ subgroups $\{A_{j}\}_{j\in[M]}$,
    we define the problem of a KL-RS PCA projection into d-dimensions with 
    target $T$ as
    \begin{equation*}
      \begin{aligned}
      &\min_{U\in\mathbb{R}^{m\times n}, rank(U)\leq d, \lambda\geq 0}\lambda\\
      &\text{s.t.}\lambda\log\left(\sum_{i=1}^{J}\frac{\vert A_{i}\vert}{m}
      \exp(\frac{1}{\lambda\vert A_{i}\vert}loss(A_{i}, A_{i}UU^{T}))\right)\leq \tau
      \end{aligned}
    \end{equation*}
\end{definition}

To some extent, our KL-RS PCA aligns with Rawls's \textit{difference principle}. While we allow for different performance levels across subgroups, our optimization goal is to minimize the disparity among all subgroups as much as possible.

\subsubsection{Experiment Details}
\textbf{Experiment platform.} The experiment platform is a Ubuntu Server 18.04.01 with 32G RAM an Intel(R) Xeon(R) W-2140B CPU @ 3,20 GHz, which has 8 cores and 16 threads. The GPU is GeForce GTX 1080 Ti and CuDA version is 11.4. 

\subsection{Hyperparameters Tuning}
\label{sec: hyperparameter tuning}

Compared to the radius $r$ in DRO and the tilted factor $\lambda$ in 
$\hat{R}(\bt, \lambda)$, the target parameter $\tau$ has a more interpretable physical meaning. 

%The target parameter $\tau$ is a natural quality which can be linked to other quantities in the optimization problem. In other words, $\tau$ can be set as a function of certain quantities, and adjusted by tuning coefficients within the function.

It is common to normalize $\tau$ by $\mathbb{E}_{\Pemp}[l(\bt_{N}^*, \tbz)]$ where we can let 
$\tau=a\mathbb{E}_{\Pemp}[l(\bt_{N}^*, \tbz)]$ with $a\geq 1$ \cite{long_robust_2023}. Then we tune the proportionality factor $a$ as opposed to directly tune $\tau$. However, this method only considers the feasibility of $\tau$, ignoring the loss spread information. In fact, we do not know what magnitude of $a$ qualifies as "sufficiently large". For example, when $\max_{i\in[N]}l(\bz_i, \bt_{N}^*)-\min_{i\in[N]}l(\bz_i, \bt_{N}^*)\leq 0.001\mathbb{E}_{\Pemp}[l(\bt_{N}^*, \tbz)]$, recall that $\bt_{N}^*$ is the optimal solution of \eqref{Prob: eop}, even $a=1.01$ should been seen as a large value.

To address this issue, we can introduce some loss spread
information into our function to adjust target performance $\tau$. One possible choice could be $\tau=a\max_{i\in[N]}l( \bt_{N}^*, \bz_i)
+(1-a)\min_{i\in[N]}l(\bt_{N}^*, \bz_i)$ where $0\leq a\leq 1$. When 
we $a$ is larger, $\tau$ is larger and we will obtain a more robust model.
It is worth noting that we need to ensure that $a$ is large enough to
ensure $\tau\geq \mathbb{E}_{\Pemp}[l(\bt_{N}^*, \tbz)]$ to ensure that Problem \eqref{Prob: klrs} has a feasible solution. Inspired by Proposition \ref{prop: mv}, another possible choice could be $\tau=\mathbb{E}_{\Pemp}[l(\bt_{N}^*, \tbz)]+a\mathbb{V}_{\Pemp}[l(\tbz, \bt_{N}^*)]$ where $a\geq 0$. Since $\tau$ increase with $a$, we will obtain a more robust model when we increase $a$. Compared to the previously mentioned approach, this method ensures that the selected $\tau$ always ensure feasiblity for Problem \eqref{Prob: klrs}. 

When applying KL-RS in practice, we can combine the aforementioned 
approach with cross-validation to select the parameters.

\subsubsection{Comparison between RS and relatives}

\begin{table}[H]
\begin{center}
    \caption{Comparison between RS and relatives.}
    \label{tab: Advantages}
    \begin{tabular}{ |l|l|l|l|l|l| } 
        \hline
        {Method} & Input & Solution & Solving & Parameter & Distribution \\
        & & Space & Algorithm & Calibration & Shift \\
        \hline
        DRO & $r$ & Subset & Specific & Hard & Part \\ 
        TREM & $\lambda$ & Subset & Specific & Hard & Standard \\ 
        RS & $\tau$ & Standard & General & Easy & Standard \\ 
        \hline
    \end{tabular}
\end{center}
\end{table}

We summarize the advantages over DRO and TERM (regularized DRO) in Table \ref{tab: Advantages}, and further elaborate on each key point in detail.

(1) Tangible Input: compared to $r$ and $\lambda$, $\tau$ is more interpretable. Model complexity is explicitly tuned via the trade-off between setting a looser target empirical risk while achieving lower loss under adversarial distributions. The selection of $\tau$ can be guided by the loss of ERM $\tau_0$. For example, $\tau$ is often expressed as $\tau_0(1+\epsilon)$, where $\underline{l}$ is the minimal empirical loss and the scalar satisfies $\epsilon\geq 0$, or $r\underline{l} + (1-r)\bar{l}$, where $\bar{l}$ is the maximum empirical loss and the scalar satisfies $r\in[0,1]$.

(2) Parameter calibration: The target parameter is inherently adaptive to problem instances, and we can easily prescribe a small range of values to search from. Existing literature has drawn some observations that calibrating the target parameter via cross-validation is easier than calibrating the radius parameter in DRO models in operations management applications. In our work, we hope to explore this in the context of neural networks and also compare with models using TERM.

(3) Solution Space: As demonstrated in (cite), the solution space of the DRO family is a subset of that of RS. Under the assumption of convex loss, the solution spaces of the DRO family and RS are identical. Given that this work addresses general machine learning problems without assuming specific loss functions, RS is the more appropriate framework.

(4) Solving Algorithm: The RS framework does not prescribe specific solving algorithms. In other words, one can choose any preferred stochastic optimization algorithm while maintaining the same computational complexity as that of ERM. However, the specific solving algorithm of TREM have three disadvantages:

Under stronger assumptions, the convergence rate is similar to SGD, which does not the benefit from acceleration or adaptive step size.
The convergence rate is exponential with the scale of $\frac{1}{\lambda}$.
The gradient is often unstable in practice because of the biasedness and scaling discussed in our Appendix B.

\section{Technical Results and Proofs}
\label{app: proof}
\subsection{Proof of Theorem \ref{thm: eq} and Proposition \ref{prop: optDistn}}
\begin{proof}\label{prof: eq}
  Problem \eqref{Prob: klrs} is equivalent to the following formulation:
  \begin{equation*}
      \begin{aligned}
          \inf_{\bt\in\Theta, \lambda\geq 0}& \quad \lambda\\
          \text{s.t.}\quad & \sup_{\mathbb{P}\ll\hat{\mathbb{P}}}\{
            \mathbb{E}_{\mathbb{P}}[l(\bt, \tbz)]-\lambda D_{KL}(\mathbb{P}
            \Vert \hat{\mathbb{P}})\}\leq \tau,   \quad \forall \bP \in 
            \{\mathbb{P}_0 \in \Pset(\Omega): \mathbb{P}_0\ll
          \Pemp\}.
  \end{aligned}
\end{equation*}
\cite{follmer2002convex, shapiro2021lectures, 
  follmer2011stochastic} provide dual formulation for the left hand 
  expression of the constraint. For convenience of further analysis , 
  we reprove the transformation as following.
  \begin{equation}\label{dual}
    \begin{aligned}
    &\sup_{\mathbb{P}\ll\hat{\mathbb{P}}}\{\mathbb{E}_{\mathbb{P}}[l(\tbz,  
    \bt)]-\lambda D_{KL}(\mathbb{P}\Vert \hat{\mathbb{P}})\}\\
    =&\sup_{\mathbb{P}\ll\Pemp}\{\mathbb{E}_{\mathbb{P}}[l(\bt, \tbz)]-\lambda\mathbb{E}_{\mathbb{P}}[\log(\frac{d\mathbb{P}}{d\Pemp})]\}\\
    =&\sup_{\mathbb{P}\ll\Pemp}\{\mathbb{E}_{\mathbb{P}}[l(\bt, \tbz)-\lambda\mathbb{E}_{\Pemp}\phi(\frac{d\mathbb{P}}{d\Pemp})]\}\\
  =&\inf_{\eta}\sup_{\mathbb{P}}\{\mathbb{E}_{\mathbb{P}}[l(f(\bt, \tbz))]-\lambda
  \mathbb{E}_{\mathbb{P}}[\phi(\frac{d\mathbb{P}}{d\Pemp})]
  +\eta-\eta\mathbb{E}_{\mathbb{P}}[1]\}\\
  =&\inf_{\eta}\{\eta+\mathbb{E}_{\hat{\mathbb{P}}}\sup_{\mathbb{P}}\{l(\bt, \tbz)
  \frac{d\mathbb{P}}{d\mathbb{\hat{\mathbb{P}}}}-\eta\frac{d\mathbb{P}}
  {d\mathbb{\hat{\mathbb{P}}}}-\lambda\frac{\mathbb{P}}{\Pemp}\phi(\frac{d\mathbb{P}}{d\hat{\mathbb{P}}})\}\}\\
  =&\inf_{\eta}\{\eta+\mathbb{E}_{\hat{\mathbb{P}}}\left[(\lambda\phi)^{\ast}
    (l(\bt, \tbz)-\eta)\right]\}\\
  =&\inf_{\eta}\{\eta+\lambda\mathbb{E}_{\hat{\mathbb{P}}}\left[\phi^{\ast}
  (\frac{l(\bt, \tbz)-\eta}{\lambda})\right]\},
    \end{aligned}
  \end{equation}

  where $\eta$ is dual variable for constraint $\mathbb{E}_{\mathbb{P}}[1]=1$ and $\phi(t)=t\log(t)-t+1$. $\phi^{\ast}(s)$ is the conjugate function of 
  $\phi(t)$, where $\phi^{\ast}(s)\triangleq\displaystyle{\sup_{t}}\{t^{\top}s-\phi(t)\}=
  \exp(s)-1$ and the corresponding optimizer is $t^{\ast}=\exp(s)$. $(\lambda\phi)^{\ast}=\lambda\phi^{\ast}(s/\lambda)$ is a classical property of conjugate function \cite{boyd2004convex}.
  So the supremum is obtained at the worst case distribution $\mathbb{P}^{\ast}(\tbz=\hbz_{n})=
  \exp(\frac{l(\bt, \hbz_{n})-\eta}{\lambda})\hat{\mathbb{P}}(\tbz=\hbz_{n})$.
  
  According to \cite{follmer2011stochastic}, the minimizer of 
  the last expression  of \eqref{dual} is $\eta^{\ast}= \lambda\log(
    \mathbb{E}_{\hat{
  \mathbb{P}}}\left[\exp\left(l(\bt, \tbz)/\lambda\right)
  \right])$, which can be 
  obtained by setting the deviation to 0. Take $\eta^{\ast}$ into the 
  expression, we can conclude  the following result 
  \begin{equation}\label{tiledlossequivalence}
  \sup_{\mathbb{P}\ll\hat{\mathbb{P}}}\{\mathbb{E}_{\mathbb{P}}[
  l(\tbz, \bt)]-\lambda D_{KL}(\mathbb{P}\Vert \hat{\mathbb{P}})\}
  =\lambda\log\left(\mathbb{E}_{\hat{\mathbb{P}}}\left[\exp\left(
  l(\bt, \tbz)/\lambda\right)\right]\right)=\hat{R}(\bt, \lambda).
  \end{equation}

  The corresponding optimal distribution $\mathbb{P}^{\ast}(\tbz=\hbz_{n})=
  \frac{\exp(l(\bt, \hbz_{n})/\lambda)}{\mathbb{E}_{\Pemp}[\exp(l(\bt, \tbz)/\lambda)]}
  \Pemp(\tbz=\hbz_{n})$.
\end{proof}

\subsection{Proof of Proposition \ref{prop: mv}}
\begin{proof}
With \eqref{tiledlossequivalence}, we can begin our discussion from $\hat{R}(\bt, \lambda)$'s sup problem formulation. 
\begin{equation*}
  \begin{aligned}
    &\displaystyle\sup_{\mathbb{P}\ll\hat{\mathbb{P}}}\left\{
      \mathbb{E}_{\mathbb{P}}[l(\bt, \tbz)]-D_{KL}
    (\mathbb{P}\Vert \hat{\mathbb{P}}) \right\}\\
    =&\displaystyle\sup_{\mathbb{P}\ll\hat{\mathbb{P}}}
    \left\{\mathbb{E}_{\mathbb{P}}[l(\bt, \tbz)]-D_{KL}
    (\mathbb{P}\Vert \hat{\mathbb{P}})-\mathbb{E}_{\mathbb{P}}
    [\mathbb{E}_{\hat{\mathbb{P}}}[l(\bt, \tbz)]] \right\}+\mathbb{E}_{\hat{\mathbb{P}}}[l(\bt, \tbz)]\\
    =&\displaystyle\inf_{\eta}\sup_{\mathbb{P}}\left\{ \eta+
    \mathbb{E}_{\hat{\mathbb{P}}}[\frac{d\mathbb{P}}{d
    \hat{\mathbb{P}}}(l(\bt, \tbz)-\eta-\mathbb{E}_{\hat{\mathbb{P}}}
    [f(\bt, \tbz)])-\lambda\phi(\frac{d\mathbb{P}}{d\hat{\mathbb{P}}})]
    \right\}+\mathbb{E}_{\hat{\mathbb{P}}}[l(\bt, \tbz)]\\
    =&\displaystyle\inf_{\eta}\left\{ \eta+\lambda\mathbb{E}_{
      \hat{\mathbb{P}}}[\phi^{\ast}(\frac{l(\bt, \tbz)-
      \eta-\mathbb{E}_{\Pemp}[l(\bt, \tbz)]}{\lambda})]\right\}+\mathbb{E}_{\hat{\mathbb{P}}}[l(\bt, \tbz)]
\end{aligned}
\end{equation*}
The first equality holds due to the fact that $\mathbb{E}_{\hat{\mathbb{P}}}
[l(\bt, \tbz)]$ is a constant. The second equality holds due to the reason
that it is a Lagrange dual expression where $\eta$ is dual variable. 
The third equality holds due to $\phi^{\ast}(s)=\sup_{x}\{x^{\top}s-\phi(s)\}=\exp(s)-1$. The Taylor's expansion of $\phi^{\ast}(s)$ around 
0 is $\phi^{\ast}(s)= s+\frac{1}{2}s^{2}+o(s^{2})$.
For a very large positive $\lambda$, we have: 
\begin{equation*}
    \begin{aligned}
        &\displaystyle\inf_{\eta}\left\{ \eta+\lambda\mathbb{E}_{\hat{\mathbb{P}}}[\phi^{\ast}(\frac{l(\bt, \tbz)-\eta-\mathbb{E}_{\Pemp}[l(\bt, \tbz)]}{\lambda})]\right\}\\
        =&\displaystyle\inf_{\eta}\left\{\eta+\lambda\mathbb{E}_{\Pemp}[(\frac{l(\bt, \tbz)-\eta-\mathbb{E}_{\Pemp}[l(\bt, \tbz)]}{\lambda})]+\frac{1}{2\lambda}\mathbb{E}_{\Pemp}[(l(\bt, \tbz)-\eta-\mathbb{E}_{\Pemp}[l(\bt, \tbz)])^{2}]+o(\frac{1}{\lambda^{2}})\right\}\\
        =&\inf_{\eta}\left\{\frac{1}{2\lambda}\mathbb{E}_{\Pemp}[l(\bt, \tbz)-\eta-\mathbb{E}_{\Pemp}[l(\bt, \tbz)]]^{2}\right\}+o(\frac{1}{\lambda^{2}})\\
        \geq&\frac{1}{2\lambda}\mathbb{E}_{\hat{\mathbb{E}}}\left[l(\bt, \tbz)-\mathbb{E}_{\hat{\mathbb{P}}}[l(\bt, \tbz)]\right]^{2}+o(\frac{1}{\lambda^{2}})\\
        =&\frac{1}{2\lambda}\mathbb{V}_{\hat{\mathbb{P}}}[l(\bt, \tbz)]+o(\frac{1}{\lambda^{2}})
      \end{aligned}
\end{equation*}

The first equality is obtained by Taylor's expansion. The inequality holds because
$\mathbb{E}_{\hat{\mathbb{P}}}[l(\bt, \tbz)]=\arg\min_{x}
\mathbb{E}_{\hat{\mathbb{P}}}[l(\bt, \tbz)-x]^{2}$ then the 
minimum is achieved when $\eta=0$.
\end{proof}

Above discussion demonstrates the relationship between $\hR$ and mean-variance when $\lambda$ is sufficient large. When we set our target $\tau$ as a number which is a little larger than the minimal value 
of Empirical Risk Minimization, a feasible $\lambda$ to problem 
\eqref{Prob: klrs} must be a large number. We discuss the problem under the following two assumptions below.

These two assumptions are not too strict assumptions. The finite mean and finite variance assumptions are commonly made. For finite samples, as long as the loss on any single sample is not infinite, both of these assumptions hold. Our paper discusses the nontrivial case, where the variance can not be zero. Because this situation effectively eliminates randomness. Without loss of generality, we can set a lower bound $\epsilon$ for the variance. Suppose the minimizer of
\eqref{Prob: eop} is $\hat{\bt}_{N}$.

\begin{proof}
We can further transform \eqref{Prob: klrs Sim} into the following form:
\begin{equation*}
    \begin{aligned}
        \inf_{\lambda\geq 0, \bt\in\Theta}& \quad \lambda\\
        \text{s.t.}\quad & \mathbb{E}_{\hat{\mathbb{P}}}\left[\exp\left(\frac{l(\bt, \tbz)-\tau}
        {\lambda}\right)\right]\leq 1.
    \end{aligned}
\end{equation*}
  In Assumption \ref{be} and Assumption \ref{bv}, we have assume that the 
  first order moment and second order moment is bounded, i.e. 
  $\Ex_{\Pemp}[l(\bt, \tbz)]\leq M_{1}$ and $\Vx_{\Pemp}[l(\bt, \tbz)]\leq M_{2}$. 
  Since $b >1$, we have $\tau\leq M_{1}+M_{2}$. Then $(l(\bt, \tbz)-\tau)/\lambda
  \geq-(M_{1}+M_{2})/\lambda$ holds. So we only discuss
  the property of $\exp(x)$ within the  interval $[-(M_{1}+M_{2})/\lambda,
  +\infty)$. Within this  interval, $\exp(x)$ function is
  $\exp(-(M_{1}+M_{2})/\lambda)$-strongly convex for $\nabla^{2}\exp(x)\geq 
  \exp(-(M_{1}+M_{2})/\lambda)$ within this interval. With the property of strongly 
  convex, we have $\exp(x)\geq 1+x+\frac{\exp(-(M_{1}+M_{2})/\lambda)}{2}x^{2}$.
  Then any feasible solution $(\bt, \lambda)$ must satisfy the following:
  
  \begin{equation*}
    \begin{aligned}
      1\geq& \mathbb{E}_{\hat{\mathbb{P}}}\left[\exp(\frac{l( 
      \bt, \tbz)-\tau}{\lambda})\right]\\
      \geq& \mathbb{E}_{\hat{\mathbb{P}}}\left[1+\frac{l( 
      \bt, \tbz)-\tau}{\lambda}+\frac{\exp(-(M_{1}+M_{2})/\lambda)}{2}\left(
        \frac{l(\bt, \tbz)-\tau}{\lambda}\right)^{2}\right]\\
      =&1+\frac{1}{\lambda}\mathbb{E}_{\hat{\mathbb{P}}}[l(\bt, \tbz)] 
      -\frac{1}{\lambda}\mathbb{E}_{\hat{\mathbb{P}}}[l(\tbz, \hat{\bt}_{N})]
      -\frac{1}{b\lambda}\mathbb{V}_{\hat{\mathbb{P}}}[l(\tbz, \hat{\bt}_{N})]\\
      &+\frac{\exp(-(M_{1}+M_{2})/\lambda)}{2\lambda^{2}}\mathbb{E}_{\hat{\mathbb{P}}}
      [l(\bt, \tbz)-\tau]^{2}\\
      \geq& 1-\frac{1}{b\lambda}\mathbb{V}_{\hat{\mathbb{P}}}[l(\hat{\bt}_{N}, \tbz)]
      +\frac{\exp(-(M_{1}+M_{2})/\lambda)}{2\lambda^{2}}\mathbb{E}_{\hat{\mathbb{P}}}
      [l(\bt, \tbz)-\tau]^{2}\\
      \geq& 1-\frac{1}{b\lambda}\mathbb{V}_{\hat{\mathbb{P}}}[l(\hat{\bt}_{N}, \tbz)]
      +\frac{\exp(-(M_{1}+M_{2})/\lambda)}{2\lambda^{2}}\mathbb{V}_{\hat{\mathbb{P}}}[l(\bt, \tbz)].
    \end{aligned}
    \end{equation*}
    The second inequality holds due to the strongly convex property. The third inequality
    holds due to $\hat{\bt}_{N}=\arg\min_{\bt\in\Theta}\mathbb{E}_{\hat{\mathbb{P}}}
    [l(\bt, \tbz)]$. The fourth inequality holds due to $
    \mathbb{E}_{\hat{\mathbb{P}}}[l(\bt, \tbz)-\tau]^{2}\geq \mathbb{E}_{\hat{\mathbb{P}}}
    [l(\bt, \tbz)-\mathbb{E}
    _{\hat{\mathbb{P}}}[l(\bt, \tbz)]]^{2}=\mathbb{V}_{\hat{\mathbb{P}}}[l(\bt, \tbz)]$
    which is a conclusion from 
    $\mathbb{E}_{\hat{\mathbb{P}}}[l(\bt, \tbz)]=\arg\min_{x}\mathbb{E}_{\hat{\mathbb{P}}}
    [l(\bt, \tbz)-x]^{2}$.
    After rearranging the above expression, we know that a feasible $\lambda$
    must satisfy the following:
    \begin{equation}\label{23}
        \lambda \exp((M_{1}+M_{2})/\lambda)\geq \frac{b}{2}\frac{\mathbb{V}_{
        \hat{\mathbb{P}}}[l(\bt, \tbz)]}{\mathbb{V}_{
        \hat{\mathbb{P}}}[l(\hat{\bt}_{N}, \tbz)]}
        \geq \frac{b\epsilon}{2\mathbb{V}_{\hat{\mathbb{P}}}
        [l(\hat{\bt}_{N}, \tbz)]}.
    \end{equation}
    Let $\psi(\lambda)=\lambda \exp((M_{1}+M_{2})/\lambda)$, then 
    $\nabla\psi(\lambda)=\exp((M_{1}+M_{2})/\lambda)(1-\frac{(M_{1}+M_{2})
    }{\lambda})$. $\psi(\lambda)$ decreses as $\lambda$ increases
    when $\lambda<(M_{1}+M_{2})$, increases as $\lambda$ increases when 
    $(M_{1}+M_{2})$, and achieves its minimum value $\psi(M_{1}+M_{2})=e(M_{1}+M_{2})$.
    With L'Hôpital's Rule we have 
    \begin{equation*}
      \begin{aligned}
      \lim_{\lambda\to 0_{+}}\psi(\lambda)=&
      \lim_{\lambda\to0_{+}}\frac{\exp((M_{1}+M_{2})/\lambda)}
      {1/\lambda}\\
      =&\lim_{x\to+\infty}\frac{\exp((M_{1}+M_{2})x)}{x}\\
      =&\lim_{x\to+\infty}\frac{\nabla\exp((M_{1}+M_{2})x)}{\nabla x}
      =+\infty.
      \end{aligned}
    \end{equation*}
    On the other hand, 
    \begin{equation*}
    \lim_{\lambda\to+\infty}\psi(\lambda)
    \geq \lim_{\lambda\to+\infty}\lambda(1+(M_{1}+M_{2})/\lambda)=+\infty.
    \end{equation*}
    From the above discussion, $\lambda$ which makes \eqref{23} hold may
    lie in two intervals, we assume they are $(0, \lambda_{1}]$
    and $[\lambda_{2}, +\infty)$ with $\lambda_{1}<\lambda_{2}$.
    It's noted that $\lambda=M_{1}+M_{2}$ is an infeasible solution for
    \eqref{Prob: klrs Sim} since it makes \eqref{23} not hold. From \eqref{Prob: klrs} we can know that any $\lambda\leq M_{1}+M_{2}$ is also infeasible. So $\lambda$ must
    lie in $[\lambda_{2}, +\infty)$. In this case, $\lambda\geq M_{1}+M_{2}$ and
    the following expression must be hold:
    \begin{equation*}
    \lambda\exp(1)\ge \lambda\exp((M_{1}+M_{2})/\lambda)\geq 
      \frac{b\epsilon}{2\mathbb{V}_{
        \hat{\mathbb{P}}}[l(\hat{\bt}_{N}, \tbz)]}.
    \end{equation*}
    Overall, we have $\lambda\geq \frac{b\epsilon}{2e\mathbb{V}_{
        \hat{\mathbb{P}}}[l(\hat{\bt}_{N}, \tbz)]}$.
\end{proof}

\subsection{Proof of Proposition \ref{prop: probability inequality}}
\begin{proof}
  \begin{equation*}
  \begin{aligned}
    \hat{\mathbb{P}}(l(\bt, \tbz)\geq \tau+\alpha)&=
    \hat{\mathbb{P}}\left(\exp(l(\bt, \tbz)/\lambda)
    \geq \exp((\tau+\alpha)/\lambda)\right)\\
    &\leq \frac{\mathbb{E}_{\hat{\mathbb{P}}}[
      \exp(l(\bt, \tbz)/\lambda)]}{\exp((\tau+\alpha)/\lambda)}\\
    &\leq \frac{\exp(\tau/\lambda)}{\exp((\tau+\alpha)/\lambda)}=\exp(\frac{-\alpha}{\lambda}).
    \end{aligned}
  \end{equation*}
  The first inequality holds due to Markov Inequality. The second inequality 
  holds because $(\bt, \lambda)$ is a feasible solution of problem(\ref{Prob: klrs Sim}). 
\end{proof}

\subsection{Proof of Theorem \ref{thm: asymptotic discrete}}

\begin{proof}
As we have mentioned in the main text, when the true distribution $\Popt$ is a discrete distribution with finite $K$ supports, the KL divergence between $\Popt$ and its empirical distribution $\Pemp$ has the following asymptotic behavior:
\begin{equation}\label{exp: aymphi}
  2N \cdot D_{KL}(\Popt\Vert\Pemp)\to \chi^{2}_{K-1} \quad \text{as } \; N\to \infty.
\end{equation}

Based on above property, for any $r\geq 0$, we have
\begin{equation}
  \mathbb{P}^{N}(D_{KL}(\Popt\Vert\Pemp)\leq r) = 
  \chi_{K-1}^{2}(\tilde{y}\leq 2Nr) \quad \text{as }
  \; N \to \infty.
\end{equation}

Consider $(\bt_{N}^{\ast}, \lambda_{N}^{\ast})$ is the optimal solution of \eqref{Prob: klrs Sim}, we have:
\begin{equation}
\mathbb{E}_{\Popt}[l(\bt_{N}^{\ast}, \tbz)]\leq \tau+\lambda_{N}^{\ast}D_{KL}(\Popt\Vert \Pemp).
\end{equation}

Combining the above expressions, we can obtain the following 
performance guarantee for $(\bt_{N}^{\ast}, \lambda_{N}^{\ast})$:

\begin{equation}
\mathbb{P}^{N}(\mathbb{E}_{\Popt}[l(\bt_{N}^{\ast}, \tbz)]\leq \tau+\lambda_{N}^{\ast}r) \geq  \chi_{K-1}^{2}(\tilde{y}\leq 2Nr)\quad \text{as }\; N\to\infty.
\end{equation}
\end{proof}

\subsection{Proof of Theorem \ref{thm: asymptotic continuous}} \label{sec: asy conti}

We can adopt the idea in Theorem~\ref{thm: asymptotic discrete} and derive a similar bound after discretization.

\begin{proof}
  For a given $\bt_{N}^{\ast}$, as we have assumed in main text that  $l(\bt_{N}^{\ast}, \tbz)$ is a measurable continuous  function. So $l(\bt_{N}^{\ast}, \tbz)$ is a random variable with 
  bounded value and continuous distribution. We use $\tilde{u}$ to denote such random variable $l(\bt_{N}^{\ast}, \tbz)$
  and use continuous distribution $\Popt_{\tilde{u}}\in\mathcal{P}(\mathcal{U})$ to denote its distribution which is supported on $\mathcal{U}=[-C, C]$. We discretize 
  $\Popt_{\tilde{u}}$ into a discrete distribution $\Popt_{\tilde{u}, K}\in
  \mathcal{P}(\mathcal{U}_{K})$ with $K$ supports, where $\mathcal{U}_{K}=\{-C, \frac{(-K+2)C}{K}, 
  \cdots, \frac{(K-2)C}{K}\}$. Let
  $\Popt_{\tilde{u}, K, k}\triangleq\Popt_{\tilde{u}, K}(\tilde{u}=-C+\frac{2(k-1)C}{K})\triangleq\Popt_{\tilde{u}}\left(\{\tilde{u}\vert -C+\frac{2(k-1)C}{K}\leq \tilde{u}\leq-C+
  \frac{2kC}{K} \}\right), \forall k\in[K]$ denote the probability density function on $k$th support $-C+\frac{2(k-1)C}{K}$. Since $\Popt_{\tilde{u}}$ is continuous distribution,
  the overlap of interval boundary does not affect the value of $\Popt_{\tilde{u}, K}$.
  Then we have 
\begin{equation}\label{32}
  \begin{aligned}
  \mathbb{E}_{\Popt_{\tilde{u}}}[\tilde{u}]&=\int_{-C}^{C}
  \tilde{u}d\Popt_{\tilde{u}}(\tilde{u})=\sum_{k\in[K]}
  \int_{-C+\frac{2(k-1)C}{K}}^{-C+\frac{kC}{K}} \tilde{u}
  d\Popt_{\tilde{u}}(\tilde{u})\\
  &\leq \sum_{k\in[K]}\int_{-C+\frac{2(k-1)C}{K}}^{-C+\frac{2kC}{K}}
  (-C+\frac{2(k-1)C}{K}+\frac{2C}{K})d\Popt_{\tilde{u}}(\tilde{u})\\
  &=\sum_{k\in[K]}(-C+\frac{2(k-1)C}{K}+\frac{2C}{K})
  \Popt_{\tilde{u}, K, k}\\
  &=\sum_{k\in[K]}(-C+\frac{2(k-1)C}{K})\Popt_{\tilde{u}, K, k}
  +\frac{2C}{K}\sum_{k\in[K]}\Popt_{\tilde{u}, K, k}\\
  &=\mathbb{E}_{\Popt_{\tilde{u}, K}}[\tilde{u}]
  +\frac{2C}{K}.
  \end{aligned}
\end{equation}
  The last equality holds due to the definition of $\Popt_{\tilde{u}, K}$. With larger $K$, the error caused by discretization will be less.
  
  We i.i.d sample $N$ samples $\{\hat{u}_{n}\}_{n\in[N]}$ 
  from distribution $\Popt_{\tilde{u}}$. Let $\Pemp_{\tilde{u}}$ denote the empirical distribution constructed by $\{\hat{u}_{n}\}_{n\in[N]}$ where $\Pemp_{\tilde{u}}[\tilde{u}=\hat{u}_{n}]=1/N$. 
  
  We also discretize the samples. Consider $K$ intervals 
  i.e. $[-C, -C+\frac{2C}{K}), [-C+\frac{2C}{K},
  -C+\frac{4C}{K}), \cdots, [\frac{2(k-1)C}{k}, C]$. For $\tilde{u}_{n}$, we quantify it as the value of the left endpoint of its interval and use $\tilde{u}_{K, n}$ to denote this value.
  Then we can obtain $N$ samples $\{\hat{u}_{K, n}\}_{n\in[N]}$ which can be regarded as 
  being i.i.d sampled from $\Popt_{\tilde{u}, K}$. We use $N_{k}$ to denote the occurrence number of $-C+\frac{2(k-1)}{K}$ in $\{\tilde{u}_{K, n}\}_{n\in[N]}$. Let $\Pemp_{\tilde{u}, K}$ denote the empirical distribution constructed by $\{\hat{u}_{K, n}\}_{n\in[N]}$ where $\Pemp_{\tilde{u}, K, k}\triangleq\Pemp_{\tilde{u}, K}(\tilde{u}=-C+\frac{2(k-1)C}{K})\triangleq\frac{N_{k}}{N}$. Then the following inequality holds for 
  any $\lambda >0$:
  
\begin{equation}\label{exp: discloss}
  \begin{aligned}
    \lambda\log(\mathbb{E}_{\Pemp_{\tilde{u}}}[
      \exp(\tilde{u}/\lambda)])
    &=\lambda\log(\frac{1}{N}\sum_{n\in[N]}\exp(\hat{u}_{n}/\lambda))\\
    &\leq\lambda\log(\frac{1}{N}\sum_{n\in[N]}\exp(\frac{\hat{u}_{K, n}+2C/K}{\lambda}))\\
    &=\lambda\log(\frac{\exp(2C/K\lambda)}{N}\sum_{n\in[N]}\exp(\frac{\hat{u}_{K, n}}{\lambda}))\\
    &=\lambda(\frac{2C}{K\lambda}+\log(\frac{1}{N}\sum_{n\in[N]}\exp(\frac{\hat{u}_{K, n}}{\lambda})))\\
    &=\lambda\log(\mathbb{E}_{\Pemp_{\tilde{u}, K}}[\exp(\tilde{u}/\lambda)])+\frac{2C}{K}
  \end{aligned}
\end{equation}
  
  The inequality hold due to the strategy dividing intervals.
  
  When $ \lambda\log(\mathbb{E}_{\Pemp_{\tilde{u}}}[\exp(\tilde{u}/\lambda)])\leq \tau$
  holds, $\lambda\log(\mathbb{E}_{\Pemp_{\tilde{u}, K}}
  [\exp(\tilde{u}/\lambda)])\leq \tau$ also holds.
  
  Consider a non-negative number $r\geq \frac{2C}{k\lambda_{N}^\ast}$ and $\overline{r}=
  r-\frac{2C}{k\lambda_{N}^\ast}$. 
  
  Similar to \eqref{exp: aymphi}, we have:
\begin{equation}
  \mathbb{P}^{N}(D_{KL}(\Popt_{\tilde{u}, K}\Vert
  \Pemp_{\tilde{u}, K})\leq \overline{r})=\chi_{K-1}^{2}(\tilde{y}
  \leq 2N\overline{r})\quad\text{as }\;N\to \infty.
\end{equation}

Then we can obtain 
\begin{equation}
\mathbb{P}^{N}(\mathbb{E}_{\Popt_{\tilde{u}, K}}[\tilde{u}]\leq \tau+\lambda_{N}^{\ast}\overline{r}) \geq \chi_{K-1}^{2}(\tilde{y}\leq 2N\overline{r})\quad \text{as }\; N\to\infty.
\end{equation}
  
  With \eqref{exp: discloss}, we can obtain
  \begin{equation*}
      \mathbb{E}_{\Popt_{\tilde{u}}}[\tilde{u}]\leq \mathbb{E}_{\Popt_{\tilde{u}, K}}[\tilde{u}]+\frac{2C}{\lambda_{N}^{\ast}}.
  \end{equation*}

  Then we can conclude the following result:
\begin{equation}
  \mathbb{P}^{N}(\mathbb{E}_{\Popt_{\tilde{u}}}[\tilde{u}]\leq \tau +\lambda^{\ast}_{N} r)
  \geq \chi_{K-1}^{2}\left(\tilde{y}\leq 2Nr-\frac{2NC}{K\lambda_{N}^{\ast}}\right)\quad\text{as }\: N\to\infty.
\end{equation}
  
  Since the above discussion holds for any positive integer $K\geq 2$, we can further strengthen 
  our result as following:
  
\begin{equation}
  \mathbb{P}^{N}(\mathbb{E}_{\Popt_{\tilde{u}}}[\tilde{u}]\leq \tau +\lambda_{N}^{\ast} r)
  \geq \sup_{K\geq 2}\left\{\chi_{K-1}^{2}\left(\tilde{y}\leq 2Nr-\frac{2NC}{K\lambda_{N}^{\ast}}\right)\right\} \quad\text{as }\; N\to\infty.
\end{equation}
\end{proof}

\subsection{Proof of Theorem \ref{thm: finite discrete}}
The proof process is similar to the proof of Theorem \ref{thm: asymptotic discrete} but with a different statistical property.
\begin{proof}
According to \cite{canonne2023concentration}, with probability at least $1-\delta$ the following inequality holds
\begin{equation*}
D_{KL}(\Popt\Vert\Pemp^{l})\leq \mathbb{E}_{\mathbb{P}^{N}}[D_{KL}(\Popt\Vert\Pemp^{l})]+\frac{6\sqrt{K\log^{5}(4K/\delta)}+311}{N}+\frac{160K}{N^{3/2}}.
\end{equation*}

This probability denote the random event following the distribution $\mathbb{P}_{N}$. Since $(\bt_{N}^{\ast}, \lambda_{N}^{\ast})$ is a feasible solution of KL-RS model on $\Pemp^{l}$, the following inequality holds
\begin{equation*}
\mathbb{E}_{\Popt}[l(\bt_{N}^{\ast}, \tbz)]\leq \tau+\lambda_{N}^{\ast}D_{KL}(\Popt\Vert\Pemp^{l}).
\end{equation*}

For a given $N$, let $r=\mathbb{E}_{\mathbb{P}^{N}}[D_{KL}(\Popt\Vert\Pemp^{l})]+\frac{6\sqrt{K\log^{5}(4K/\delta)}+311}{N}+\frac{160K}{N^{3/2}}$. Combining the two expressions above, the following probability inequality holds
\begin{equation*}
    \mathbb{P}^{N}(\mathbb{E}_{\Popt}[l(\bt_{N}^{\ast}, \tbz)]\leq \tau+\lambda_{N}^{\ast}r)\geq 1-\delta.
\end{equation*}
\end{proof}

\subsection{Proof of Theorem \ref{thm: hierachicalquivalence}}

\begin{proof}
  To prove Theorem \ref{thm: hierachicalquivalence}, we need to utilize \eqref{tiledlossequivalence} in \ref{prof: eq}.
  \begin{equation}
    \begin{aligned}
    &\mathbb{E}_{\mathbb{P}_{\tbz, \tbg}}\left[l(\bt, \tbz)\right]
    \leq \tau+\lambda_{1}D_{KL}(\mathbb{P}_{\tbg}\Vert \Pemp_{\tbg})
    +\lambda_{2}\mathbb{E}_{\mathbb{P}_{\tbg}}D_{KL}(\mathbb{P}_{\tbz\vert\tbg}
    \Vert\Pemp_{\tbz\vert\tbg}), \forall\;\; \mathbb{P}_{\tbz, \tbg}\ll\Pemp_{\tbz, \tbg}\\
    \Longleftrightarrow&\sup_{\mathbb{P}_{\tbz\vert\tbg}\ll\Pemp_{\tbz\vert\tbg}}
    \mathbb{E}_{\mathbb{P}_{\tbg}}\left[\mathbb{E}
    _{\Pemp_{\tbz\vert 
    \tbg}}[l(\bt, \tbz)]-\lambda_{2}D_{KL}(\mathbb{P}
    _{\tbz\vert\tbg}
   \Vert\Pemp_{\tbz\vert\tbg})
    \right]\\
    &\leq \tau+\lambda_{1}D_{KL}(\mathbb{P}
    _{\tbg}\Vert\Pemp_{\tbg}), \forall\;\;  \mathbb{P}_{\tbg}\ll\Pemp_{\tbg}\\
    \Longleftrightarrow&\mathbb{E}_{\mathbb{P}_{\tbg}}\left[\lambda_{2}\log
    \left(\mathbb{E}_{\Pemp_{\tbz\vert\tbg}}
    \exp\left(f(\bt, \tbz)/\lambda_{2}\right)\right) \right]
    \leq \tau +\lambda_{1}D_{KL}(\mathbb{P}
    _{\tbg}\Vert\Pemp_{\tbg}), \forall\;\;  \mathbb{P}_{\tbg}\ll\Pemp_{\tbg}\\
    \Longleftrightarrow&\sup_{\mathbb{P}_{\tbg}\ll\Pemp_{\tbg}}
    \mathbb{E}_{\mathbb{P}_{\tbg}}\left[\lambda_{2}\log\left(
    \mathbb{E}_{\Pemp_{\tbz\vert\tbg}}
    \exp\left(l(\bt, \tbz)/\lambda_{2}\right)\right) 
    -\lambda_{1}D_{KL}(\mathbb{P}_{\tbg}\Vert\Pemp_{\tbg})\right]\leq \tau\\
    \Longleftrightarrow&\lambda_{1}\log\left(\mathbb{E}_{\Pemp_{\tbg}}
    \left[ \exp(\lambda_{2}\log\left(
      \mathbb{E}_{\Pemp_{\tbz\vert\tbg}}
      \exp\left(l(\bt, \tbz)/\lambda_{2}\right)\right)/\lambda_{1})\right]
    \right)\leq \tau
  \end{aligned}
  \end{equation}
  The first and third $\Longleftrightarrow$ are due to the inequality holding for 
  any distribution within the support.
  
  The second and third $\Longleftrightarrow$ hold for the equality \ref{tiledlossequivalence}.
\end{proof}

\subsection{Proof of Proposition \ref{prop: Lp KLRS}}
\begin{proof}
    
    Since affine operator and exponential operator preserve convexity\cite{boyd2004convex},
    when $l(\bt, \tbz)$ is convex with $\bt$, $f(\bt, \tbz;\lambda)$ is convex with $\bt$. 
    
    W.l.o.g, we can assume $0 \leq l(\bt, \tbz) \leq C, \forall \bt\in\Theta$. 
    
    When $l(\bt, \tbz)$ is $\mu_{l}$-strongly convex with $\bt$. For $0\leq b\leq 1$, we have the following 
    \begin{equation}
    \begin{aligned}
      &bf(\bt, \tbz;\lambda)+(1-b)f(\bt', \tbz;\lambda)\\
      =&b\exp(\frac{l(\bt)-\tau}{\lambda})+(1-b)\exp(\frac{l(\bt')-\tau}{\lambda})\\
      \geq&\exp(\frac{bl(\bt)+(1-b)l(\bt')-\tau}{\lambda})\\ 
      \geq&\exp(\frac{l(b\bt+(1-b)\bt')+\frac{\mu_{l}}{2}
      \Vert \bt-\bt'\Vert^{2}_{2}-\tau}{\lambda}) \\
      \geq&\exp(\frac{l(b\bt+(1-b)\bt')-\tau}{\lambda})+\exp(\frac{l(b\bt+(1-b)
      \bt')-\tau}{\lambda})(\frac{\mu_{l}}{2})\Vert \bt-\bt'\Vert^{2}_{2}\\
      \geq&\exp(\frac{l(b\bt+(1-b)\bt')-\tau}{\lambda})+\frac{\mu_{l}}{2}
      \exp(\frac{-\tau}{\lambda})
      \Vert \bt-\bt'\Vert^{2}_{2}\\
      =&f(b\bt+(1-b)\bt', \tbz; \lambda)+\frac{\mu_{l}}{2}\exp(-\tau/\lambda)
      \Vert \bt-\bt'\Vert^{2}_{2}.
    \end{aligned} 
  \end{equation}
The first inequality holds because $\exp$ function is a convexity function. The second inequality
holds because $l(\bt, \tbz)$ is $\mu_{l}$-strongly convex. The third inequality is because $\exp(x)\geq x+1$. The fourth inequality is because $l$ is non-negative.
    
 When $l(\bt, \lambda)$ is $C_{l}$-lipschitz continuous, we have 
 \begin{equation*}
\Vert \nabla f(\bt, \tbz; \lambda)\Vert=\Vert \frac{\nabla l(\bt, \tbz)}
  {\lambda}\exp\left(\frac{l(\bt, \tbz)-\tau}{\lambda}\right)\Vert
  \leq \frac{C_{l}}{\lambda}\exp\left(\frac{C-\tau}{\lambda}\right),
 \end{equation*}
which can infer $f(\bt, \tbz;\lambda)$ is also lipschitz continuous.

When $l(\bt, \lambda)$ is $S_{l}$-lipschitz smooth, we have 
\begin{equation*}
  \begin{aligned}
      &\Vert \nabla^{2}f(\bt, \tbz; \lambda)\Vert\\
      =&\exp\left(\frac{l(\bt, \tbz)-\tau}{\lambda}\right)\frac{1}{\lambda}
      \Vert \nabla^{2}l(\bt, \tbz)+\frac{\nabla l(\bt, \tbz)}{\lambda}\nabla 
      l(\bt, \tbz)^{\top}\Vert\\
      \leq&\left(\frac{C_{l}^{2}}{\lambda^{2}}+\frac{S_{l}}{\lambda}\right)
      \exp\left((C-\tau)/\lambda)\right),
  \end{aligned}
\end{equation*}
which infers that $f(\bt,\tbz;\lambda)$ is also lipschitz smooth.
\end{proof}

\section{Limitations}
\label{app: limi}
Our paper does not discuss the case of using more general $\phi$ divergences. The main reason is that when using general $\phi$ divergence, we cannot obtain a reformulation similar to \eqref{Prob: klrs Sim}. This necessitates simultaneously optimizing both the distribution and the parameter $\bt$. The exploration of more general $\phi$-divergences might be a direction for future research.

Additionally, our research makes minimal assumptions about the true distribution $\Popt$, leading to relatively general conclusions. When our KL-RS framework is applied to specific problems, the true distribution may exhibit more definite characteristics, such as a normal distribution or a Bernoulli distribution. In these cases, the expression of our problem might be further simplified and presented in a more insightful mathematical form.
Exploring whether KL-RS can yield more insightful conclusions in more specific machine learning tasks compared to general cases might be a direction for future research.

\section{Potential Society Impacts}
\label{app: impacts}
We believe our work will have a positive social impact. More researchers will pay attention to the potential applications of the RS framework, which focuses on target-oriented robust optimization, in machine learning. Especially for high-risk task scenarios, our KL-RS framework might offer robust solutions. Adjusting the KL-RS framework based on specific application scenarios will further enrich the framework's content and generate insights beyond those mentioned in this paper.

%\input{formal_paper/appendix_temp}
%\newpage
%\input{formal_paper/appendix}

%\newpage
%\input{formal_paper/checklist}

\end{document}